\documentclass[10pt,twocolumn,letterpaper]{article}

\usepackage[pagenumbers]{cvpr} 

\usepackage{graphicx}
\usepackage{tabularx}
\usepackage{siunitx}
\usepackage{booktabs}       
\usepackage{multirow}
\usepackage{multicol}
\usepackage{makecell}

\usepackage[table]{xcolor}

\definecolor{darkred}{RGB}{139,0,0}
\definecolor{darkblue}{RGB}{0,0,139}
\definecolor{lightblue}{RGB}{242,249,255}
\definecolor{lightpink}{RGB}{255,242,235}
\definecolor{lightgrey}{RGB}{242,242,242}
\definecolor{lightyellow}{RGB}{252,248,221}

\usepackage{amsmath}

\usepackage{ulem}
\usepackage{soul}
\usepackage{pifont}
\usepackage{amssymb}

\definecolor{cvprblue}{rgb}{0.21,0.49,0.74}
\usepackage[pagebackref,breaklinks,colorlinks,allcolors=cvprblue]{hyperref}

\def\paperID{11105} 
\def\confName{CVPR}
\def\confYear{2026}

\title{OmniZip: Learning a Unified and Lightweight Lossless Compressor for Multi-Modal Data}

\author{%
  Yan Zhao\textsuperscript{\rm 1}, 
  Zhengxue Cheng\textsuperscript{\rm 1}\thanks{Corresponding author.}, Junxuan Zhang\textsuperscript{\rm 2}, 
  Dajiang Zhou\textsuperscript{\rm 2}, 
  Qunshan Gu\textsuperscript{\rm 2}, Qi Wang\textsuperscript{\rm 2}, 
  Li Song\textsuperscript{\rm 1}\protect\footnotemark[1] \\
  \textsuperscript{\rm 1}Shanghai Jiao Tong University\quad
  \textsuperscript{\rm 2}Ant Group\\
  \textsuperscript{\rm 1}\{zhaoyanzy, zxcheng, song\_li\}@sjtu.edu.cn\\
  \textsuperscript{\rm 2}\{junxuan.zjx, dajiang.zdj, qunshan.gu, qw.qq\}@antgroup.com
}

\begin{document}
\maketitle
\begin{abstract}

Lossless compression is essential for efficient data storage and transmission. Although learning-based lossless compressors achieve strong results, most of them are designed for a single modality, leading to redundant compressor deployments in multi-modal settings. Designing a unified multi-modal compressor is critical yet challenging, as different data types vary largely in format, dimension, and statistics. Multi-modal large language models offer a promising resolution but remain too complex for practical use. Thus, we propose \textbf{OmniZip}, \textbf{a unified and lightweight lossless compressor for multi-modal data (like image, text, speech, tactile, database, and gene sequence)}. Built on a lightweight backbone, OmniZip incorporates three key components to enable efficient multi-modal lossless compression: a modality-unified tokenizer that reversibly transforms diverse data into tokens, a modality-routing context learning mechanism that enables flexible multi-modal context modeling, and a modality-routing feedforward design that further enhances the model's nonlinear representation flexibility. A reparameterization training strategy is used to enhance model capacity. OmniZip outperforms or matches other state-of-the-art compressors on multiple modalities, achieving 42\%, 57\%, 62\% and 42\%, 53\% higher compression efficiency than gzip on CLIC-M, TouchandGo, enwik9, LibriSpeech, and WikiSQL datasets, respectively. It also supports near real-time inference on resource-constrained edge devices, reaching about 1MB/s on MacBook CPUs and iPhone NPUs. Our code is released at \url{https://github.com/adminasmi/OmniZip-CVPR2026}.

\end{abstract}    
\vspace{-3pt}
\section{Introduction}
\label{sec:intro}

According to information theory \cite{codingtheory}, the optimal code length for a symbol is determined by its negative logarithmic probability. This insight motivates the modern lossless compression strategy: use probabilistic models to estimate data likelihoods and apply entropy coding \cite{Huffman:1952, AC:1991, ANS:2013} to encode data. Recent approaches further leverage neural networks and large language models (LLMs) to improve probability estimation and enhance compression performance \cite{lmic,llmzip,tszip,p2llm,l3tc,msdzip,l3c,Heurtel,aiwave,bcm,junhao,dualcomp,taccompress,taco}.



Despite recent progress, learning-based lossless compression faces two key challenges. 
\begin{itemize}
    \item \textit{First}, \textit{many existing methods, especially those based on LLMs} \cite{llmzip, junhao, p2llm, lmic}\textit{, suffer from excessive complexity.} These models often contain billions of parameters, far exceeding the size of compressed data. Since compression speed is decided by the model's inference speed, such complexity severely limits these methods' practical deployments. For example, compressing a single 1080p image using LLaMA3-8B \cite{llama3} takes over 30 minutes \cite{junhao, p2llm}, while \cite{nncpv2:2021, cmix:2023} need several days to compress 1 GB of text.   
    \item {\textit{Second, most existing methods are designed for a single modality} \cite{llmzip, dlpr, p2llm, l3tc}, \textit{necessitating separate compressors in multi-modal systems.}} This increases software complexity and hardware costs. Despite its importance, multi-modal lossless compression remains challenging due to the heterogeneity of modalities:
    text is discrete and sequential, images are two-dimensional and spatially organized, speech is continuous with smooth spectral patterns, database contains structured categorical fields, and gene sequences exhibit symbolic structures with characteristic motifs. Some works \cite{lmic} explore multi-modal lossless compression by converting all data to ASCII text, causing sub-optimal performance on non-text modalities.
\end{itemize}

\begin{figure*}[!tb]
    \centering
    \includegraphics[width=\linewidth]{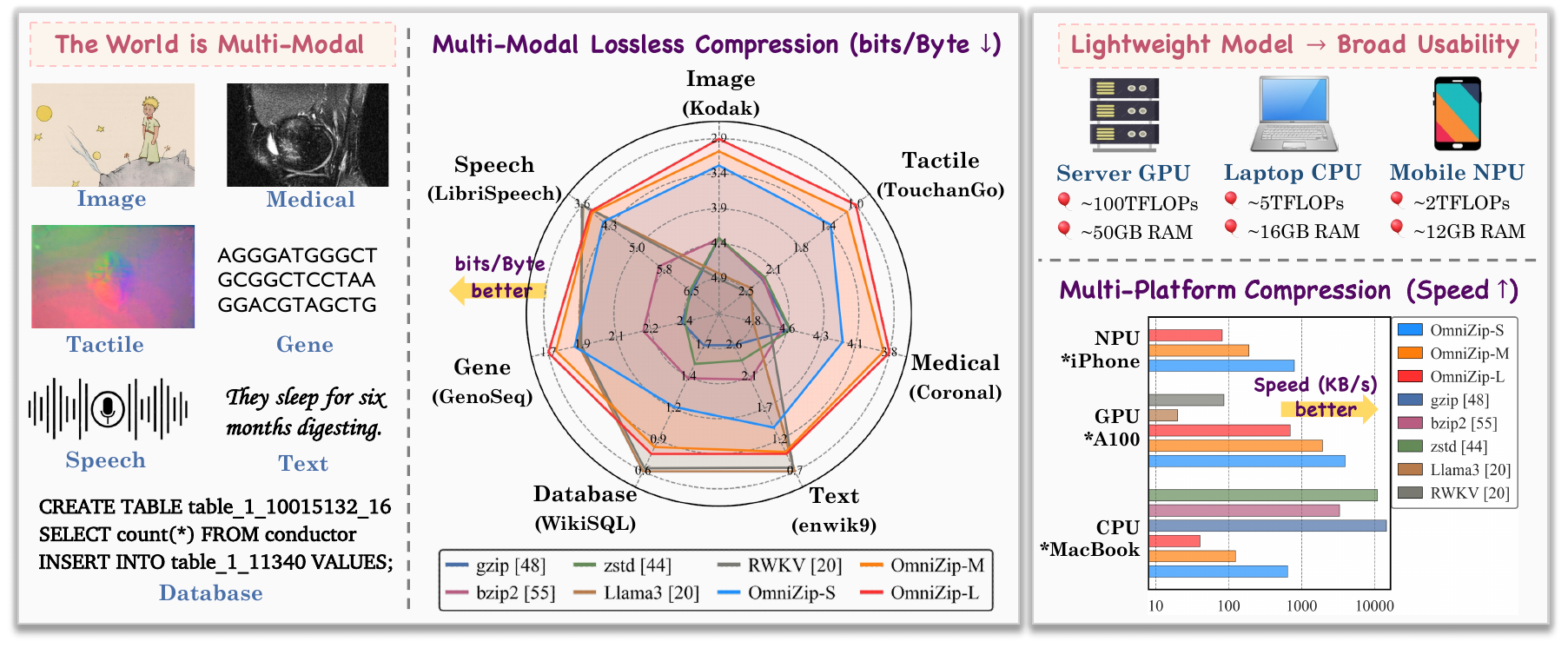}
    \vspace{-24pt}
    \caption{\textbf{Left}: The world is multi-modal, motivating the need for unified lossless compression. Multi-modal compressors' performance is shown across modalities, where points closer to the edge indicate better efficiency (lower bits/Byte). \textbf{Right}: Lightweight design ensures broad usability across platforms. OmniZip achieves near real-time inference (hundreds of KB/s to MB/s) even on edge devices.}
    \label{fig:teaser}
    \vspace{-11pt}
\end{figure*}

To tackle these challenges, we propose \textit{\textbf{OmniZip}}, a unified and lightweight lossless compressor for multi-modal data. {It supports most real-world data types suitable for lossless compression: image-like data (e.g., natural image, medical image, tactile signals), text-like data (e.g., natural language, gene sequence, database), and speech.} Built upon a lightweight backbone, OmniZip explicitly addresses modality heterogeneity and enables efficient multi-modal compression through three key designs: 
\textit{\textbf{(1) a modality-unified tokenizer}} that maps diverse data into a unified and fully reversible token space, 
\textit{\textbf{(2) a modality-routing context learning mechanism}} that enables flexible multi-modal context modeling, 
and \textit{\textbf{(3) a modality-routing feedforward process}} that introduces adaptive feedforward representations. 
A \textit{\textbf{reparameterization training strategy}} further enhances the model capacity without increasing the inference cost. 

With much fewer parameters, OmniZip matches or surpasses the SOTA compressors across diverse modalities. It supports deployment across CPUs, NPUs, and GPUs, achieving near real-time inference even on resource-constrained edge devices (e.g., laptops and mobile phones).



Generally, our contributions can be concluded as:


\begin{itemize}
    \item We propose OmniZip, a unified and lightweight lossless compressor for multi-modal data. It supports efficient lossless compression for image-like (natural image, medical image, tactile signal), text-like (natural language, gene sequence, database), and speech data. OmniZip explicitly addresses modality heterogeneity and provides useful insights for the emerging multi-modal applications.
    
    

    \item Built on a lightweight backbone, OmniZip integrates three key techniques for efficient multi-modality lossless compression: modality-unified tokenization, modality-routing contextual learning, and modality-routing feedforward. A reparameterization training strategy is also employed to enhance the model's capacity.

    \item We evaluate OmniZip on 16 datasets across seven modalities. It outperforms or matches other SOTA compressors, achieving 42\%, 57\%, 62\% and 42\%, 53\% higher compression efficiency than gzip on CLIC-M, TouchandGo, enwik9, LibriSpeech, and WikiSQL datasets. Even on edge devices like MacBook CPU or iPhone NPU, OmniZip reaches near real-time inference (0.1$\sim$1MB/s).
    
    


\end{itemize}
\section{Related Work}

\paragraph{Classical Lossless Compressors.}

Classical lossless compression can be broadly categorized into general-purpose and modality-specific approaches. General-purpose compressors such as gzip \cite{gzip}, bzip2 \cite{bzip2}, and zstd \cite{zstd} operate across diverse data types without domain-specific assumptions. Gzip integrates LZ77 \cite{lz77:1977} and Huffman coding \cite{Huffman:1952}, bzip2 builds on the Burrows-Wheeler transform \cite{burrows-wheeler} and move-to-front encoding \cite{movetofront}, while zstd uses fast LZ-style matching along with finite-state entropy coding \cite{finitecoding}.

Modality-specific compressors, in contrast, leverage domain-specific statistical structures for higher efficiency. For images, codecs like PNG \cite{png}, WebP \cite{webp}, FLIF \cite{flif}, JPEG XL \cite{jpegxl}, JPEG 2000 \cite{jpeg2000}, and BPG \cite{bpg} exploit spatial redundancy with filtering, prediction, or transform coding. For lossless speech compression, FLAC \cite{flac} uses linear predictive coding to model waveforms and compresses the residual signals with entropy coding.



\vspace{-11pt}
\paragraph{Learning-based Lossless Compressors.}

Learning-based compression has progressed rapidly in both lossy \cite{cheng2020,xujun,fengdh,taccomp,tcm,funcodec} and lossless domains \cite{dlpr,young,l3c,l3tc,aiwave,bcm,lmic,Heurtel}. We focus on lossless compression, which combines probability prediction with entropy coding to achieve bit-exact reconstruction. 
These methods are typically designed for a single modality, as the statistical properties of images, text, and speech differ greatly, making cross-modality generalization challenging. Existing text compressors use auto-regressive architectures to model token sequences \cite{finezip, llmzip, nncpv2:2021, cmix:2023, l3tc, tszip}, while image compressors attempt to capture the spatial dependencies \cite{junhao, p2llm, l3c, dlpr, ivpf, iflow, rc}. While \cite{lmic,Heurtel} explore multi-modal lossless compression, they treat all modalities as byte or ASCII text, overlooking the modality heterogeneity and leading to sub-optimal performance. In addition, learning-based compressors, especially those built upon LLMs \cite{junhao, p2llm, finezip, llmzip, tszip, lmic}, often suffer from heavy computational complexity.



\section{Proposed Method}

\begin{figure}[!tb]
    \centering
    \includegraphics[width=\linewidth]{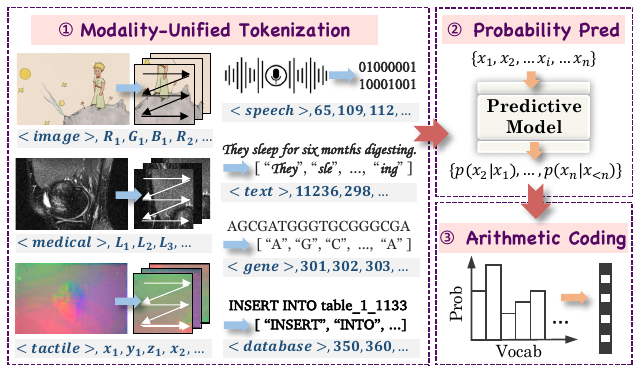}
    \vspace{-20pt}
    \caption{Overview of the proposed OmniZip framework. Diverse data is first converted into a unified, fully reversible token space. A predictive model then estimates each token’s contextual probability, followed by arithmetic coding to generate the bitstream.}
    \label{fig:overview}
    \vspace{-12pt}
\end{figure}

As outlined in \cref{fig:overview}, OmniZip enables lossless compression for multiple modalities, supporting image-like data (natural image, medical image, tactile signal), text-like data (natural language, gene sequence, database), and speech. Input data is first converted into tokens $\{x_1, x_2, ..., x_n\}$ through a modality-unified tokenizer. A predictive model estimates each token's contextual probability $p(x_i|x_{<i})$. Then, arithmetic coding (AC) \cite{AC:1991} encodes the data towards its entropy bound $H(p) = \mathop{\mathbb{E}} (\sum_{i=1}^{n} -\log_{2} p(x_i|x_{<i})])$. 

\subsection{Preliminaries: Low-Complexity Backbone}

\paragraph{Parameter-Efficient Prediction Model.}
\label{sec:choose}

To lay a foundation for efficient compression, we compare three common backbones of prediction models: Transformer \cite{transformer:2017}, Mamba \cite{mamba}, and RWKV \cite{rwkv7}. They are evaluated based on compression performance (bits/Byte), computational cost (MACs), and inference speed (KB/s) on a MacBook CPU. For simplicity, we focus on text compression for model selection and do not consider multi-modal data.


Specifically, text data is tokenized using SentencePiece BPE tokenizer \cite{bpe} (vocabulary size is 16K), and then split into token sequences of length 1024 for model input. The three models perform auto-regressive next-roken probability prediction, with arithmetic coding applied to generate the compressed bitstream. All models are trained on enwik8 \cite{enwik8} and evaluated on enwik9 \cite{enwik9}. As shown in \cref{tab:archcomp}, under such lightweight model settings, RWKV achieves the best balance of compression efficiency and inference speed, making it the preferred backbone. The Mamba model, after adaptation for CPU compatibility, exhibits relatively slow inference and is thus excluded from further comparisons.




\vspace{-10pt}
\paragraph{Reparameterization Training Strategy.}
\label{sec:hira}

\begin{table}[!tb]
    \centering
    \footnotesize
    \renewcommand{\arraystretch}{0.98}
    \setlength{\tabcolsep}{2.8pt}
    \caption{Comparison for model backbone selection. Models are evaluated using text lossless compression. Inference speed is measured on the MacBook Pro CPU, with a batch size of 128.}
    \vspace{-8pt}
    \begin{tabular}{lcccc}
    \toprule
     \textbf{Model} 
    & \textbf{\#Params} $\downarrow$
    & \textbf{MACs} $\downarrow$
    & \textbf{bits/Byte} $\downarrow$ 
    & \textbf{\makecell{Speed (KB/s)}} $\uparrow$\\
    \midrule
    \multirow{2}{*}{Transformer \cite{transformer:2017}} 
    & 0.2M & 0.88M & 2.197 & 714 \\
    & 3.2M & 4.43M & 1.984 & 280 \\
    \midrule
    \multirow{2}{*}{RWKV-7 \cite{rwkv7}} 
    & 0.2M & 0.85M  & 1.910 & 2292 \\
    & 3.2M & 3.68M  & 1.658 & 856 \\
    \bottomrule
    \end{tabular}
    \vspace{-12pt}
    \label{tab:archcomp}
\end{table}

Inspired by \cite{l3tc}, we employ a reparameterization strategy \cite{reparam} to enhance the model's performance without affecting inference speed. During training, additional branches are added to increase the model's capacity. These branches are then merged back into the main path during inference, ensuring that inference complexity remains unchanged. Following \cite{l3tc}, we apply reparameterization to the receptance (R), key (K), and value (V) layers in the Time Mixing module of RWKV. To further increase the training capacity, we apply high-rank matrix decomposition to the added branches \cite{l3tc}.



\subsection{Modality-Unified Tokenization}
\label{sec:tokenization}


To enable lossless compression across diverse modalities, we design a unified tokenization strategy that can reversibly represent diverse data within a unified token space. The key challenge lies in the structural and statistical differences among modalities: for instance, images are two-dimensional with 8-bit RGB sub-pixels, text is discrete and highly variable, and speech is continuous with smooth temporal patterns. Furthermore, the tokenization must remain fully invertible to ensure lossless compression.


Specifically, we group data into three categories: image-like data, text-like data, and speech. For text-like data (natural language, gene, and database), we follow \cite{l3tc} and use a SentencePiece BPE tokenizer with a 16K vocabulary. To further improve efficiency, domain-specific symbols are explicitly added to the vocabulary, including nucleotide bases (A, T, G, C) for gene sequences and common script keywords (e.g., "select", "or", "and") for databases.

For image-like data, including natural images, medical images, and tactile signals, we first divide each image into $16\times16\times3$ patches to preserve local spatial correlations. Within each patch, pixels are flattened in a raster-scan order, and each pixel’s RGB values are sequentially expanded as sub-pixels ($R_1,G_1,B_1,R_2,G_2,B_2,\dots$). Each sub-pixel is treated as an individual token, yielding a vocabulary of size 256. Tactile images captured by visual sensors are processed like natural images, while tactile force data are mapped from 3D $(x, y, z)$ signals into RGB images for tokenization. For medical images that are typically grayscale, each pixel intensity is treated as a single token.


For speech data, which is inherently continuous and difficult to discretize without loss, we directly read the raw byte stream and treat each byte as a token, also producing a 256-size vocabulary. By merging each modality's vocabulary, we create a unified token set for probability prediction. To help the model better distinguish between different modalities, we prepend each token sequence with a modality-specific prefix, i.e., $<$image$>$, $<$medical$>$, $<$tactile$>$, $<$text$>$, $<$gene$>$, $<$database$>$, or $<$speech$>$. 

Before softmax and arithmetic coding, modality masking is applied to zero out probabilities of non-target modalities, thereby reducing estimation errors and enhancing compression efficiency. For instance, when compressing images, only image-related token probabilities are retained.



\subsection{Modality-Routing Context Learning}
\label{sec:moa}



In our RWKV backbone, each block consists of two components: a Time Mixing module \cite{rwkv7} and a multilayer perceptron (MLP) \cite{mlp}. The Time Mixing module models contextual dependencies through operations like token shift, R/K/V projection, WKV calculation, and state updates.


As shown in \cref{fig:overview}, different modalities exhibit distinct contextual dependencies. To enhance multi-modal processing, we integrate a mixture-of-experts (MoE) mechanism \cite{moe} into the Time Mixing module. Considering computational efficiency, we experiment with applying MoE to different components and find that applying MoE to the V layer yields impressive results. This phenomenon may stem from the inherent mechanism that, while the K layer serves as a semantic index and the R layer functions as a memory gating, the V layer represents the specific memory content. The diversity within the V layer is more crucial in handling multi-modal data. Therefore, we apply MoE only to the V layer, with the K and R layers shared across all V experts. 

\begin{figure}
    \centering
    \includegraphics[width=\linewidth]{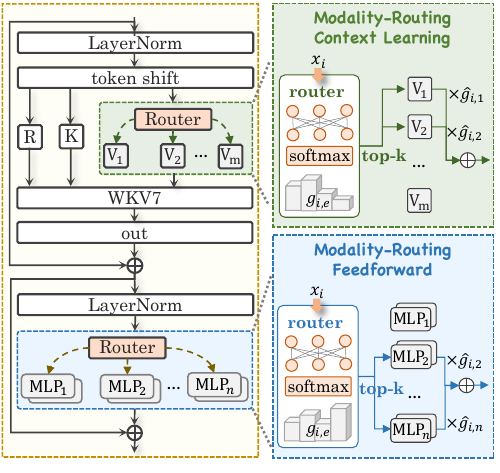}
    \vspace{-22pt}
    \caption{One block of OmniZip’s predictive model. The model stacks $N$ such blocks in total. Built on a lightweight RWKV7 backbone, it incorporates two modality-routing MoE modules for contextual learning and feedforward processing.}
    \label{fig:structure}
    \vspace{-12pt}
\end{figure}

Specifically, as shown in \cref{fig:structure}, a learnable router assigns each token $x_i$ a score $g_{i,e}$ for each expert $e$: 
\vspace{-6pt}
\begin{equation}
\small
g_{i,e} = \text{softmax}(x_i W_g)_e = \frac{\exp(x_i W_{g,e})}{\sum_{e'=1}^{E} \exp(x_i W_{g,e'})},
\vspace{-6pt}
\end{equation}
where $W_g$ is the routing projection and $E$ is the number of experts. Experts with the top-$k$ scores then process the token. Herein, each expert is a V projection layer, and the router is a small feedforward network. The final output is the weighted sum of the selected experts’ outputs: 
\vspace{-7pt}
\begin{equation}
\small
    \text{V}(x_i) = \sum_{e\in \text{top}-k}\hat{g}_{i,e}\cdot e(x_i)
\vspace{-7pt}
\end{equation}
where $e(x_i)$ denotes the output of expert $e$ for token $x_i$, and $\hat{g}{i,e}$ is the re-normalized score. To ensure a compact model design, we experimentally use 4 experts, with top-$k$ set to 2. This only adds the computational and parameter overhead of 3 additional V layers per Time Mixing module, which is negligible compared to the entire model.

\subsection{Modality-Routing Feedforward Process}
\label{sec:moe}

In RWKV, each block integrates a feedforward MLP to enhance the nonlinear representations \cite{rwkv7}. However, using a general-purpose MLP is not well-suited for multi-modal processing, as it lacks the flexibility to effectively capture the distinct characteristics of different modalities. 

To better support multi-modal compression, we replace it with an MoE-based modality-routing feedforward module, as shown in \cref{fig:structure}. The routing strategy follows \cref{sec:moa}, except that the experts here are small MLPs instead of V projection layers. Each MLP expert uses a hidden factor of $2\times$, half that of the original large MLP. With four experts and top-$k=2$, the number of activated parameters per token is roughly the same as the original MLP, while the MoE design offers greater flexibility for multi-modal processing.

\subsection{Loss Function}

Our loss function consists of two parts: (1) a primary cross-entropy loss that measures the divergence between the predicted and target distributions, and (2) auxiliary regularization terms that stabilize mixture-of-experts training, including the router Z-loss \cite{zloss} and the load-balancing loss \cite{auxloss}.




\vspace{-13pt}
\begin{equation}
\footnotesize
\begin{aligned}
\mathcal{L} &=
\underbrace{-\sum q \log p}_{\text{Cross Entropy}} \\
&+ \lambda
\underbrace{
\frac{1}{T} \sum_{j=1}^{T}
\left(
\log \sum_{t=1}^{N} e^{x_{i,t}}
\right)^{2}
}_{\text{Z-loss}} \\
&+ \mu
\Big[
\underbrace{\mathrm{CV}^2\!\left(\sum g_{i,e}\right)}_{\text{Expert Importance}}
+
\underbrace{\mathrm{CV}^2\!\left(\sum \mathbb{I}(g_{i,e}>0)\right)}_{\text{Load Balance}}
\Big]
\end{aligned}
\vspace{-2pt}
\label{equ:loss}
\end{equation}

As shown in \cref{equ:loss}, the Z-loss penalizes large logits in the gating network, preventing instability during routing and accelerating convergence. The auxiliary loss encourages balanced expert utilization by penalizing the variance in expert importance and token assignment, where $\mathrm{CV}^2$ denotes the squared coefficient of variation \cite{cv2}, 
and $\mathbb{I}(g_{i,e}>0)$ is the indicator of expert selection. The loss terms are weighted using $\lambda=0.001$ and $\mu=0.01$.

\section{Experiments}

\subsection{Experimental Setup}

\paragraph{Implementation Details.}

We vary the model size by adjusting the embedding dimension and the number of blocks, creating three model variants: \textbf{\textit{OmniZip-S}}, \textbf{\textit{OmniZip-M}}, and \textbf{\textit{OmniZip-L}}, as shown in \cref{tab:model-details}. 

\begin{table}[h]
    \vspace{-7pt}
    \centering
    \footnotesize
    \renewcommand{\arraystretch}{0.98}
    \setlength{\tabcolsep}{7.5pt}
    \caption{Details of the proposed OmniZip models.}
    \vspace{-9pt}
    \begin{tabular}{lcccc}
    \toprule
     \textbf{Model} 
    & \textbf{Blocks}
    & \textbf{Embed Dim}
    & \textbf{\#Params}
    & \textbf{MACs}\\
    \midrule
    \textbf{OmniZip-S} & 2 & 320 &  4.8M & 3.88M\\
    \textbf{OmniZip-M} & 2 & 912 &  38M  & 18.2M\\
    \textbf{OmniZip-L} & 3 & 1488 & 152M & 56.0M\\
    \bottomrule
    \end{tabular}
    \label{tab:model-details}
    \vspace{-8pt}
\end{table}

The models are trained using the FusedAdam optimizer \cite{fusedadam} with a \textit{three-stage training strategy} to stabilize the optimization of mixture-of-experts: (1) freeze the feedforward routing module and train the remaining parameters for two epochs with a learning rate of $1\times10^{-4}$; (2) freeze the contextual routing module and train the other components for another two epochs with a learning rate of $1\times10^{-4}$; (3) unfreeze all parameters and train the entire model for 20 epochs using a cosine annealing learning rate scheduler \cite{scheduler}, starting at $1\times10^{-4}$ and decaying to $1\times10^{-5}$. All training process is conducted on NVIDIA A100 GPUs. 

\begin{table*}[tb]
    \belowrulesep=0pt
    \aboverulesep=0pt
    \centering
    \renewcommand{\arraystretch}{1.0}
    \setlength{\tabcolsep}{1.7pt}
    \newcommand{\centerdash}[1]{\ifx#1-\multicolumn{1}{c}{-}\else#1\fi}
    \caption{Lossless compression results on image-like (natrual image, medical image, and tactile) datasets. Multi-modal compressors (except ours) are shown in gray. $*$ denotes pretrained LLMs, $\dag$ indicates that some results are reprodeced by us. $\prime$ indicates compression speeds measured on a MacBook CPU, and $\ddag$ denotes speeds on an A100 GPU (batch size 128). Other unmarked values are taken from their papers.} 
    \scriptsize   
    \vspace{-9pt}
    \begin{tabular}
    {p{0.18cm}p{1.65cm}|p{1.09cm}ll|c|p{1.09cm}p{1.09cm}p{1.09cm}p{1.09cm}|p{1.33cm}p{1.33cm}|p{1.08cm}p{1.08cm}p{1.08cm}}
    \toprule
     & \multirow{2}{*}{\textbf{Compressor}} 
     & \multirow{2}{*}{\textbf{\#Params}$\downarrow$} 
     & \multirow{2}{*}{\textbf{MACs}$\downarrow$} 
     & \multirow{2}{*}{\makecell{\textbf{\makecell{Speed\\(KB/s)}}$\uparrow$}}
     & \multirow{2}{*}{\textbf{\makecell{Multi\\Modal}}} 
     & \multicolumn{9}{c}{\textbf{bits/Byte$\downarrow$}}\\[1pt]
     \cmidrule{7-15}

     & & & & & 
     & \textbf{Kodak} & \textbf{CLIC-P} & \textbf{CLIC-M} & \textbf{DIV2K} 
     & \textbf{TouchandGo} & \textbf{ObjectFolder} 
     & \textbf{Axial} & \textbf{Coronal} & \textbf{Sagittal}\\
    \midrule
    \multirow{9.5}{*}{\rotatebox{90}{{\textbf{Classical}}}} 
    & \cellcolor{lightgrey}gzip$^\dag$ \cite{gzip}   
    & \cellcolor{lightgrey}- 
    & \cellcolor{lightgrey}- 
    & \cellcolor{lightgrey}14500$^\prime$ 
    & \cellcolor{lightgrey}\checkmark 
    & \cellcolor{lightgrey}4.349$_{\textbf{(0\%)}}$ 
    & \cellcolor{lightgrey}4.039$_{\textbf{(0\%)}}$
    & \cellcolor{lightgrey}3.947$_{\textbf{(0\%)}}$
    & \cellcolor{lightgrey}4.224$_{\textbf{(0\%)}}$
    & \cellcolor{lightgrey}2.298$_{\textbf{(0\%)}}$
    & \cellcolor{lightgrey}3.969$_{\textbf{(0\%)}}$
    & \cellcolor{lightgrey}5.346$_{\textbf{(0\%)}}$
    & \cellcolor{lightgrey}4.563$_{\textbf{(0\%)}}$
    & \cellcolor{lightgrey}5.571$_{\textbf{(0\%)}}$\\
    
    & \cellcolor{lightgrey}bzip2$^\dag$ \cite{bzip2} 
    & \cellcolor{lightgrey}- 
    & \cellcolor{lightgrey}- 
    & \cellcolor{lightgrey}3300$^\prime$ 
    & \cellcolor{lightgrey}\checkmark
    & \cellcolor{lightgrey}4.359 
    & \cellcolor{lightgrey}4.033 
    & \cellcolor{lightgrey}3.931 
    & \cellcolor{lightgrey}4.208 
    & \cellcolor{lightgrey}2.288 
    & \cellcolor{lightgrey}4.031 
    & \cellcolor{lightgrey}5.403 
    & \cellcolor{lightgrey}4.617 
    & \cellcolor{lightgrey}5.628\\
    
    & \cellcolor{lightgrey}zstd$^\dag$ \cite{zstd}   
    & \cellcolor{lightgrey}- 
    & \cellcolor{lightgrey}- 
    & \cellcolor{lightgrey}11000$^\prime$ 
    & \cellcolor{lightgrey}\checkmark
    & \cellcolor{lightgrey}4.350 
    & \cellcolor{lightgrey}4.042 
    & \cellcolor{lightgrey}3.952 
    & \cellcolor{lightgrey}4.229 
    & \cellcolor{lightgrey}2.263 
    & \cellcolor{lightgrey}3.966 
    & \cellcolor{lightgrey}5.343 
    & \cellcolor{lightgrey}4.560 
    & \cellcolor{lightgrey}5.568\\

    & PNG$^\dag$ \cite{png}     & - & - & 200$^\prime$ 
    & \ding{55} 
    & 4.349 
    & 4.041
    & 3.952
    & 4.229
    & 2.500
    & 3.964
    & 5.361
    & 4.581
    & 5.580\\
    & FLIF$^\dag$ \cite{flif}   & - & - & 652$^\prime$ 
    & \ding{55} 
    & 2.903 & 2.792 & 2.497 & 2.914 
    & 0.808 & 3.765 
    & 4.860 & 3.999 & 5.160\\
    & BPG$^\dag$ \cite{bpg}     & - & - & 180$^\prime$ 
    & \ding{55} 
    & 3.382 & 3.144 & 2.855 & 3.285 
    & 0.936 & 3.726 
    & 5.004 & 4.167 & 5.337\\
    & WebP$^\dag$ \cite{webp}   & - & - & 330$^\prime$ 
    & \ding{55} 
    & 3.205 & 3.006 & 2.774 & 3.176 
    & 0.739 & 3.612 
    & 4.944 & 4.161 & 5.214\\
    & JPEG-XL$^\dag$ \cite{jpegxl}  & - & - & 970$^\prime$ & \ding{55} 
    & 2.902 & 2.722 & 2.395 & 2.826 
    & 0.739 & 3.657 
    & 4.719 & 3.891 & 5.091\\
    & JPEG2000$^\dag$ \cite{jpeg2000} & - & - & 5000$^\prime$ & \ding{55} 
    & 3.301 & 2.927 & 2.667 & 3.127 
    & 1.552 & 3.989 
    & 5.019 & 4.170 & 5.310\\
    
    \midrule

    \multirow{8.5}{*}{\rotatebox{90}{\textbf{Learning-based}}}
    & \cellcolor{lightgrey}$\text{Llama3}^{*\dag}$ \cite{lmic} 
    & \cellcolor{lightgrey}8B 
    & \cellcolor{lightgrey}7.80G 
    & \cellcolor{lightgrey}20$^\ddag$ 
    & \cellcolor{lightgrey}\checkmark 
    & \cellcolor{lightgrey}4.862 
    & \cellcolor{lightgrey}4.290 
    & \cellcolor{lightgrey}4.234 
    & \cellcolor{lightgrey}4.378 
    & \cellcolor{lightgrey}2.455 
    & \cellcolor{lightgrey}4.060 
    & \cellcolor{lightgrey}5.514
    & \cellcolor{lightgrey}4.832
    & \cellcolor{lightgrey}5.784\\
    
    & \cellcolor{lightgrey}$\text{RWKV}^{*\dag}$ \cite{lmic}  
    & \cellcolor{lightgrey}7B 
    & \cellcolor{lightgrey}7.19G  
    & \cellcolor{lightgrey}86$^\ddag$ 
    & \cellcolor{lightgrey}\checkmark 
    & \cellcolor{lightgrey}4.937 
    & \cellcolor{lightgrey}4.310 
    & \cellcolor{lightgrey}4.298 
    & \cellcolor{lightgrey}4.718 
    & \cellcolor{lightgrey}2.523 
    & \cellcolor{lightgrey}4.118 
    & \cellcolor{lightgrey}5.244 
    & \cellcolor{lightgrey}4.712 
    & \cellcolor{lightgrey}5.529\\
    
    & DLPR$^{*\dag}$ \cite{dlpr}   & 22.3M & - & 640  & \ding{55} 
    & 2.860  & 2.380 & 2.160 & 2.550 
    & 1.082 & 3.774 
    & - & - & -\\
    
    & L3C$^{*\dag}$ \cite{l3c}     & 5M & 6.86M & 178 & \ding{55} & 3.260  & 2.940 & 2.640 & 3.090 
    & 1.350 & 3.842 
    & 5.160 & 4.450 & 5.520\\
    
    & RC \cite{rc}       & - & - & - & \ding{55} 
    & - & 2.930 & 2.540 & 3.080 
    & - & - & - & - & -\\
    
    & P2LLM \cite{p2llm} & 8B & - & 20$^\ddag$ & \ding{55} 
    & 2.830
    & 2.350 
    & 2.080 
    & 2.510 
    & - & - & - & - & -\\

    & aiWave \cite{aiwave} & 695M & - & 40 & \ding{55}
    & - & - & - & -
    & - & - 
    & 4.550 & 3.800 & 4.830\\
    & BCM \cite{bcm} & 12.5M & - & - & \ding{55}
    & - & - & - & -
    & - & - 
    & 4.410 & 3.630 & 4.810\\
    
    \midrule

    \multirow{3.2}{*}{\rotatebox{90}{\textbf{Ours}}}
    & \cellcolor{lightpink}OmniZip-S   
    & \cellcolor{lightpink}4.8M 
    & \cellcolor{lightpink}3.88M
    & \cellcolor{lightpink}1223$^\ddag$  
    & \cellcolor{lightpink}\checkmark 
    & \cellcolor{lightpink}3.307$_{\textbf{(-24\%)}}$  
    & \cellcolor{lightpink}2.933$_{\textbf{(-27\%)}}$
    & \cellcolor{lightpink}2.620$_{\textbf{(-34\%)}}$
    & \cellcolor{lightpink}2.968$_{\textbf{(-30\%)}}$ 
    & \cellcolor{lightpink}1.338$_{\textbf{(-42\%)}}$
    & \cellcolor{lightpink}3.275$_{\textbf{(-18\%)}}$
    & \cellcolor{lightpink}4.560$_{\textbf{(-15\%)}}$
    & \cellcolor{lightpink}4.179$_{\textbf{(-8\%)}}$
    & \cellcolor{lightpink}5.286$_{\textbf{(-5\%)}}$\\
    
    & \cellcolor{lightpink}OmniZip-M   
    & \cellcolor{lightpink}38M 
    & \cellcolor{lightpink}18.2M
    & \cellcolor{lightpink}1015$^\ddag$  
    & \cellcolor{lightpink}\checkmark 
    & \cellcolor{lightpink}3.103$_{\textbf{(-29\%)}}$  
    & \cellcolor{lightpink}2.926$_{\textbf{(-28\%)}}$
    & \cellcolor{lightpink}2.378$_{\textbf{(-40\%)}}$
    & \cellcolor{lightpink}2.749$_{\textbf{(-35\%)}}$
    & \cellcolor{lightpink}1.110$_{\textbf{(-52\%)}}$
    & \cellcolor{lightpink}2.973$_{\textbf{(-25\%)}}$
    & \cellcolor{lightpink}4.531$_{\textbf{(-15\%)}}$
    & \cellcolor{lightpink}3.880$_{\textbf{(-15\%)}}$
    & \cellcolor{lightpink}4.969$_{\textbf{(-11\%)}}$\\
    
    & \cellcolor{lightpink}OmniZip-L   
    & \cellcolor{lightpink}152M 
    & \cellcolor{lightpink}56.0M
    & \cellcolor{lightpink}420$^\ddag$  
    & \cellcolor{lightpink}\checkmark 
    & \cellcolor{lightpink}2.925$_{\textbf{(-33\%)}}$  
    & \cellcolor{lightpink}2.578$_{\textbf{(-36\%)}}$
    & \cellcolor{lightpink}2.273$_{\textbf{(-42\%)}}$
    & \cellcolor{lightpink}2.542$_{\textbf{(-40\%)}}$
    & \cellcolor{lightpink}0.987$_{\textbf{(-57\%)}}$
    & \cellcolor{lightpink}2.689$_{\textbf{(-32\%)}}$
    & \cellcolor{lightpink}4.466$_{\textbf{(-16\%)}}$
    & \cellcolor{lightpink}3.837$_{\textbf{(-16\%)}}$
    & \cellcolor{lightpink}4.848$_{\textbf{(-13\%)}}$\\
    
    \bottomrule
    \end{tabular}
    \vspace{-5pt}
    \label{tab:compare-image-tactile-medical}
\end{table*}

\begin{table*}[tb]
    \belowrulesep=0pt
    \aboverulesep=0pt
    \centering
    \renewcommand{\arraystretch}{1.0}
    \setlength{\tabcolsep}{2.7pt}
    \newcommand{\centerdash}[1]{\ifx#1-\multicolumn{1}{c}{-}\else#1\fi}

    \caption{Lossless compression results on text-like (natural text, gene sequence, database) and speech datasets. Multi-modal compressors (except ours) are shown in gray. $*$ denotes pretrained LLMs, $\dag$ indicates that some results are reprodeced by us. $\prime$ indicates compression speeds measured on a MacBook CPU, and $\ddag$ denotes speeds on an A100 GPU (batch size 128). Other values are taken from their papers.}
    \scriptsize   
    \vspace{-9pt}
    \begin{tabular}
    {p{0.25cm}p{1.7cm}|p{1.25cm}ll|c|p{1.25cm}p{1.25cm}|p{1.25cm}p{1.25cm}|p{1.25cm}p{1.5cm}|p{1.5cm}}
    \toprule
     & \multirow{2}{*}{\textbf{Compressor}} 
     & \multirow{2}{*}{\textbf{\#Params}$\downarrow$} 
     & \multirow{2}{*}{\textbf{MACs}$\downarrow$} 
     & \multirow{2}{*}{\makecell{\textbf{\makecell{Speed\\(KB/s)}}$\uparrow$}}
     & \multirow{2}{*}{\textbf{\makecell{Multi\\Modal}}} 
     & \multicolumn{7}{c}{\textbf{bits/Byte$\downarrow$}}\\[1pt]
     \cmidrule{7-13}

     & & & & & 
     & \textbf{enwik9} & \textbf{Gutenberg} & \textbf{Spider} & \textbf{WikiSQL} 
     & \textbf{GenoSeq} & \textbf{DNACorpus} 
     & \textbf{LibriSpeech}\\
    \midrule
    \multirow{4.5}{*}{\rotatebox{90}{{\textbf{Classical}}}} 
    & \cellcolor{lightgrey}gzip$^\dag$ \cite{gzip}   
    & \cellcolor{lightgrey}- 
    & \cellcolor{lightgrey}- 
    & \cellcolor{lightgrey}14500$^\prime$ 
    & \cellcolor{lightgrey}\checkmark 
    & \cellcolor{lightgrey}2.590$_{\textbf{(0\%)}}$ 
    & \cellcolor{lightgrey}3.819$_{\textbf{(0\%)}}$ 
    & \cellcolor{lightgrey}2.289$_{\textbf{(0\%)}}$ 
    & \cellcolor{lightgrey}1.672$_{\textbf{(0\%)}}$ 
    & \cellcolor{lightgrey}2.390$_{\textbf{(0\%)}}$ 
    & \cellcolor{lightgrey}2.318$_{\textbf{(0\%)}}$ 
    & \cellcolor{lightgrey}6.511$_{\textbf{(0\%)}}$\\
    
    & \cellcolor{lightgrey}bzip2$^\dag$ \cite{bzip2} 
    & \cellcolor{lightgrey}- 
    & \cellcolor{lightgrey}- 
    & \cellcolor{lightgrey}3300$^\prime$ 
    & \cellcolor{lightgrey}\checkmark
    & \cellcolor{lightgrey}2.082 
    & \cellcolor{lightgrey}3.607 
    & \cellcolor{lightgrey}2.326 
    & \cellcolor{lightgrey}1.403 
    & \cellcolor{lightgrey}2.202 
    & \cellcolor{lightgrey}2.148 
    & \cellcolor{lightgrey}5.647\\
    
    & \cellcolor{lightgrey}zstd$^\dag$ \cite{zstd}   
    & \cellcolor{lightgrey}- 
    & \cellcolor{lightgrey}- 
    & \cellcolor{lightgrey}11000$^\prime$ 
    & \cellcolor{lightgrey}\checkmark
    & \cellcolor{lightgrey}2.364 
    & \cellcolor{lightgrey}3.703
    & \cellcolor{lightgrey}2.258
    & \cellcolor{lightgrey}1.520 
    & \cellcolor{lightgrey}2.404 
    & \cellcolor{lightgrey}2.324 
    & \cellcolor{lightgrey}6.467 \\

    & $\text{FLAC}^{\dag}$ \cite{flac} & - & - & 2000$^\prime$  
    & \ding{55}
    & - & - 
    & - & -
    & - & - 
    & 4.961\\
    
    \midrule

    \multirow{6.8}{*}{\rotatebox{90}{\textbf{Learning-based}}}
    & \cellcolor{lightgrey}$\text{Llama3}^{*\dag}$ \cite{lmic} 
    & \cellcolor{lightgrey}8B 
    & \cellcolor{lightgrey}7.80G 
    & \cellcolor{lightgrey}20$^\ddag$ 
    & \cellcolor{lightgrey}\checkmark 
    & \cellcolor{lightgrey}0.722 
    & \cellcolor{lightgrey}1.030 
    & \cellcolor{lightgrey}0.770 
    & \cellcolor{lightgrey}0.645 
    & \cellcolor{lightgrey}1.889 
    & \cellcolor{lightgrey}1.808 
    & \cellcolor{lightgrey}3.616\\
    
    & \cellcolor{lightgrey}$\text{RWKV}^{*\dag}$ \cite{lmic}  
    & \cellcolor{lightgrey}7B 
    & \cellcolor{lightgrey}7.19G  
    & \cellcolor{lightgrey}86$^\ddag$ 
    & \cellcolor{lightgrey}\checkmark 
    & \cellcolor{lightgrey}0.774 
    & \cellcolor{lightgrey}1.036 
    & \cellcolor{lightgrey}0.723 
    & \cellcolor{lightgrey}0.669 
    & \cellcolor{lightgrey}1.900 
    & \cellcolor{lightgrey}1.890 
    & \cellcolor{lightgrey}3.589\\

    & $\text{tszip}^{\dag}$ \cite{tszip} & 169M & 131M & 180 & \ding{55}
    & 1.083 & 1.169 
    & 0.768 & 0.967
    & 1.969 & 1.965 
    & -\\
    & $\text{NNCP}^{\dag}$ \cite{nncpv2:2021} & - & 187M & 1.6 & \ding{55}
    & 0.853 & 1.320 
    & 1.033 & 0.700
    & 1.732 & 1.748
    & -\\
    & $\text{CMIX}^{\dag}$ \cite{cmix:2023} & - & - & 4 & \ding{55}
    & 0.879 & 1.363
    & 1.056 & 0.629 
    & 1.656 & 1.667 
    & -\\
    & $\text{L3TC}^{\dag}$ \cite{l3tc} & 12M & 13.0M & 580 & \ding{55}
    & 1.281 & 2.237
    & 2.270 & 1.712 
    & 2.368 & 2.346 
    & -\\
    \midrule

    \multirow{3.2}{*}{\rotatebox{90}{\textbf{Ours}}}
    & \cellcolor{lightpink}OmniZip-S   
    & \cellcolor{lightpink}4.8M 
    & \cellcolor{lightpink}3.88M 
    & \cellcolor{lightpink}1223$^\ddag$ 
    & \cellcolor{lightpink}\checkmark 
    & \cellcolor{lightpink}1.370$_{\textbf{(-47\%)}}$ 
    & \cellcolor{lightpink}1.552$_{\textbf{(-60\%)}}$
    & \cellcolor{lightpink}1.320$_{\textbf{(-42\%)}}$
    & \cellcolor{lightpink}1.170$_{\textbf{(-30\%)}}$
    & \cellcolor{lightpink}1.868$_{\textbf{(-24\%)}}$
    & \cellcolor{lightpink}2.015$_{\textbf{(-22\%)}}$
    & \cellcolor{lightpink}4.155$_{\textbf{(-36\%)}}$\\
    
    & \cellcolor{lightpink}OmniZip-M   
    & \cellcolor{lightpink}38M 
    & \cellcolor{lightpink}18.2M
    & \cellcolor{lightpink}1015$^\ddag$ 
    & \cellcolor{lightpink}\checkmark 
    & \cellcolor{lightpink}1.009$_{\textbf{(-61\%)}}$  
    & \cellcolor{lightpink}1.278$_{\textbf{(-67\%)}}$
    & \cellcolor{lightpink}1.298$_{\textbf{(-43\%)}}$
    & \cellcolor{lightpink}0.842$_{\textbf{(-50\%)}}$
    & \cellcolor{lightpink}1.777$_{\textbf{(-26\%)}}$
    & \cellcolor{lightpink}1.810$_{\textbf{(-22\%)}}$
    & \cellcolor{lightpink}3.852$_{\textbf{(-41\%)}}$\\
    
    & \cellcolor{lightpink}OmniZip-L   
    & \cellcolor{lightpink}152M 
    & \cellcolor{lightpink}56.0M
    & \cellcolor{lightpink}420$^\ddag$ 
    & \cellcolor{lightpink}\checkmark 
    & \cellcolor{lightpink}0.980$_{\textbf{(-62\%)}}$  
    & \cellcolor{lightpink}1.175$_{\textbf{(-69\%)}}$
    & \cellcolor{lightpink}1.243$_{\textbf{(-46\%)}}$
    & \cellcolor{lightpink}0.787$_{\textbf{(-53\%)}}$
    & \cellcolor{lightpink}1.740$_{\textbf{(-27\%)}}$
    & \cellcolor{lightpink}1.782$_{\textbf{(-23\%)}}$
    & \cellcolor{lightpink}3.810$_{\textbf{(-42\%)}}$\\
    
    \bottomrule
    \end{tabular}
    \vspace{-14pt}
    \label{tab:compare-text-gene-database-speech}
\end{table*}


\vspace{-10pt}
\paragraph{Multi-Modality Datasets.} 
Our dataset covers seven types: natural image, medical image, tactile signal, text, gene sequence, database, and speech. In total, 16 datasets are utilized for training and evaluation:
\begin{itemize}
    \item \textit{\textbf{Image:}} we train on the training set of DIV2K \cite{div2k}, and evaluate on Kodak \cite{kodak}, CLIC-Pro, CLIC-Mobile \cite{clic}, and the test set of DIV2K. 
    
    \item \textit{\textbf{Medical:}} we adopt MRNet Axial, Coronal, and Sagittal datasets \cite{mrnet} and follow their official train–test splits.
    
    \item \textit{\textbf{Text:}} the training set consists of enwik8 \cite{enwik8} and 2000 Gutenberg eBooks \cite{gutenberg}. The test set is enwik9 \cite{enwik9} and additional 1000 Gutenberg eBooks.

    \item \textit{\textbf{Speech:}} we use the LibriSpeech dataset \cite{librispeech} with its official training and test splits.
    
    \item \textit{\textbf{Database:}} we train on the Spider training set \cite{spider}, and evaluate on both WikiSQL \cite{wikisql} and the Spider test set.
    
    \item \textit{\textbf{Gene:}} we use GenoSeq \cite{genoseq} and DNACorpus \cite{dnacorpus} datasets, and follow their official train-test splits.
    
    \item \textit{\textbf{Tactile:}} the training and test sets follow the official partitions of TouchandGo \cite{touchandgo} and ObjectFolder \cite{objectfolder}.
\end{itemize}

To ensure balanced training across modalities, we control each modality's training set to 1GB via random sampling and data augmentation. Further, we also implement a balanced batch sampler to ensure even sampling from each modality in every batch by cycling through shuffled indices.





\vspace{-10pt}
\paragraph{Compared Methods and Metrics.}




We compare lossless compression across different modalities. Classical baselines include general-purpose lossless compressors (gzip \cite{gzip}, bzip2 \cite{bzip2}, zstd \cite{zstd}), image-specific compressors (PNG \cite{png}, WebP \cite{webp}, FLIF \cite{flif}, JPEG XL \cite{jpegxl}, JPEG2000 \cite{jpeg2000}, BPG \cite{bpg}), and the speech lossless compressor FLAC \cite{flac}. For learning-based methods, we evaluate recent image lossless compressors \cite{dlpr, p2llm, l3c, rc, aiwave, bcm}, text compressors \cite{nncpv2:2021, cmix:2023, l3tc, tszip}, and multi-modal compressors \cite{lmic} which compresses data using pretrained LLMs \cite{llama3, rwkv4}. Herein, image-like data (natural images, medical images, and tactile data) can be compressed using image and general-purpose compressors, while text-like data (natural language, gene, and databases) are compressed using text and general-purpose compressors. There is currently no dedicated learning-based lossless speech compressor.


We evaluate compression performance using bits/Byte as metrics. Model complexity is assessed by MACs and inference speed (KB/s) on multiple devices (an NVIDIA A100 GPU, a MacBook Pro CPU, and an iPhone 17 Pro NPU).


\subsection{Multi-Modality Lossles Results}

\begin{figure*}[!tb]
    \centering
    \includegraphics[width=\linewidth]{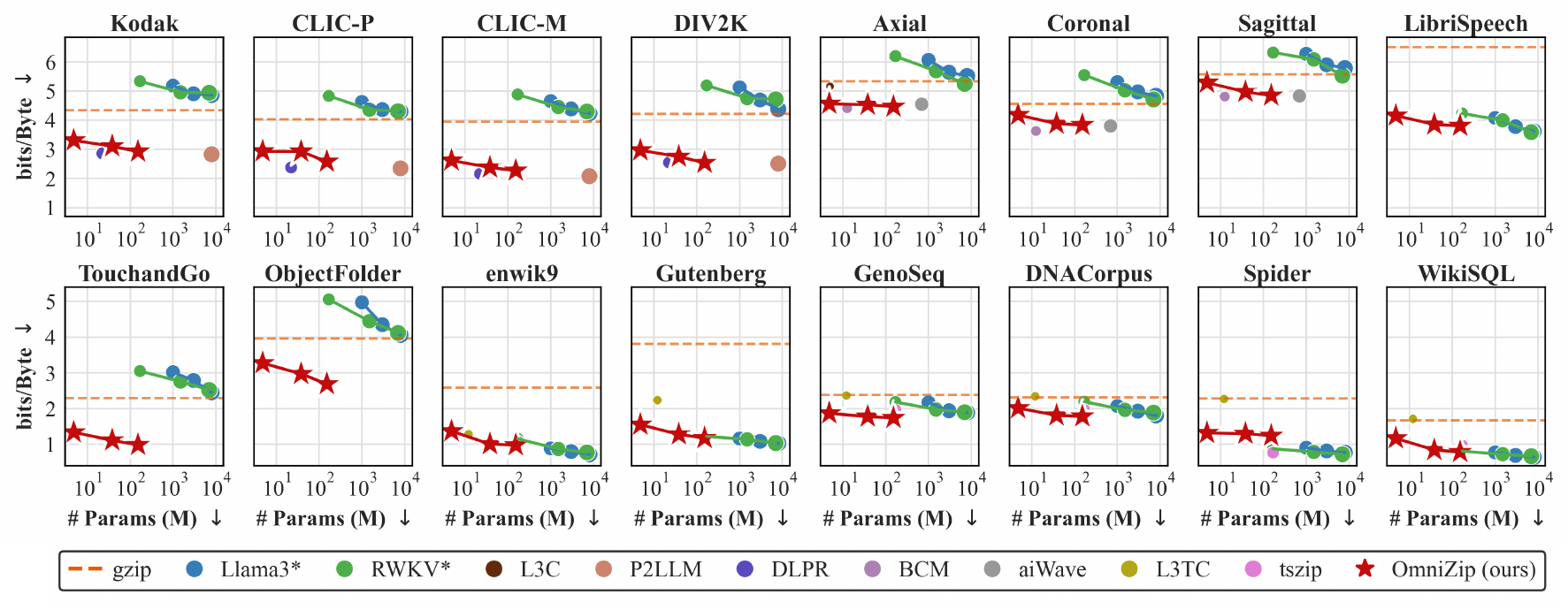}
    \vspace{-22pt}
    \caption{Comparison of learning-based lossless compressors across multi-modal datasets. The x-axis shows the model size (in millions of parameters), and the y-axis indicates compression efficiency (bits/Byte, lower is better). Models closer to the lower-left corner achieve better compression with fewer parameters. The dashed orange line represents the performance baseline of gzip.}
    \label{fig:curves}
    \vspace{-10pt}
\end{figure*}

\cref{tab:compare-image-tactile-medical} and \cref{tab:compare-text-gene-database-speech} compare OmniZip with existing lossless compression methods. The former reports results on image-like data (natural image, medical image, and tactile data), while the latter presents results on text-like data (natural text, gene sequence, database) and speech. Both tables include general-purpose multi-modal compressors (colored in gray) and modality-specific compressors for comparison. The radar chart in \cref{fig:teaser} compares multi-modal compressors across different modalities, while \cref{fig:curves} offers comparisons of learning-based compressors on individual datasets.



\vspace{-12pt}
\paragraph{Compression Results on Image-like Data.}
As shown in \cref{tab:compare-image-tactile-medical}, general-purpose classical compressors such as gzip, bzip2, and zstd show limited effectiveness on image-like data, achieving about 4.3, 2.3, 4.6 bits/Byte on Kodak, TouchandGo, and Coronal, respectively. Among classical image compressors, FLIF and JPEG-XL perform the best, offering roughly $1.5\times$$\sim$$2.5\times$ better compression than gzip.

Learning-based methods exhibit clear domain dependency. DLPR, L3C, RC, and P2LLM, all trained on natural images, perform worse on tactile and medical data. Though \citet{lmic} presents as a general-purpose compressor, it compresses images by converting them into ASCII sequences and applying pretrained LLMs \cite{llama3, rwkv4} for probability prediction, resulting in sub-optimal performance.

In contrast, OmniZip achieves pleasing compression across all image-like modalities. Even the smallest version, OmniZip-S, performs comparably to specialized learning-based methods and outperforms gzip by about 30\% on medical, tactile, and natural images. OmniZip-M and OmniZip-L further improve compression performance close to state-of-the-art levels, reaching 2.925, 0.987, and 3.937 bits/Byte on Kodak, TouchandGo, and Coronal datasets, respectively. 

\vspace{-12pt}
\paragraph{Compression Results on Text-like Data.}

\cref{tab:compare-text-gene-database-speech} presents the lossless compression results on natural language, gene sequence, and database. Classical compressors like gzip, bzip2, and zstd achieve moderate performance (about 2.3 bits/Byte on enwik9, Spider, and GenoSeq). Learning-based text compressors perform strongly on natural language data, with NNCP and CMIX reaching 0.85 and 0.88 bits/Byte on enwik9. Pretrained LLMs compress natural language more effectively, but are computationally heavy and yield only modest gains on gene sequences.

OmniZip achieves competitive compression performance on these text-like datasets with much fewer parameters. OmniZip-S achieves about 24\% to 47\% performance gains compared to gzip across all text-like datasets. OmniZip-M and OmniZip-L, further enhance compression performance by another 10\%$\sim$20\%. The performance gains on database and gene data are more pronounced, with OmniZip-L achieving near SOTA compression (1.740 and 1.782 bits/Byte) on GenoSeq and DNACorpus datasets.



\vspace{-12pt}
\paragraph{Compression Results on Speech Data.}

As shown in \cref{tab:compare-text-gene-database-speech}, classical compressors such as gzip, bzip2, and zstd achieve 5.6$\sim$6.5 bits/Byte on LibriSpeech, while the specialized speech compressor FLAC performs better at 4.96 bits/Byte. Currently, there is no dedicated learning-based lossless speech compressor. The multi-modal method \cite{lmic} provides promising compression (about 3.6 bits/Byte on LibriSpeech) but suffers from heavy complexity.

In contrast, OmniZip offers a superior balance between compression and efficiency, achieving 36\%$\sim$42\% improvements over gzip and 15$\sim$23\% over FLAC on LibriSpeech.


\begin{figure}[!tb]
    \centering
    \includegraphics[width=\linewidth]{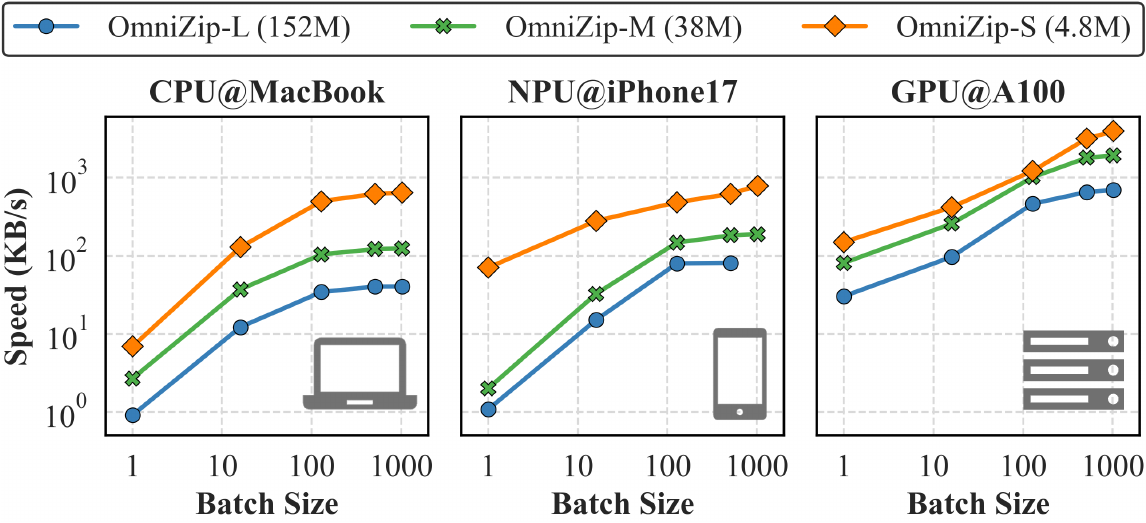}
    \vspace{-18pt}
    \caption{OmniZip's inference speed across various platforms (CPU of MacBook Pro, NPU of iPhone17 Pro, and GPU of NVIDIA A100) and batch sizes (1, 16, 128, 512, 1024).}
    \label{fig:speed}
    \vspace{-14pt}
\end{figure}

\subsection{Parameter and Complexity Analysis}

We evaluate OmniZip’s inference speed across different hardware platforms and batch sizes, as shown in \cref{fig:speed}. The inference speed improves as the batch size increases and saturates around a batch size of 512. Leveraging the modality-routing sparse design, OmniZip activates only a subset of parameters per token and achieves high computational efficiency. Specifically, on an NVIDIA A100 GPU, all three models achieve MB/s-level inference speeds, with OmniZip-S peaking at approximately 4MB/s. On resource-constrained devices like MacBook CPU and iPhone NPU, OmniComp-S achieves a near real-time speed of $\sim$1MB/s, and the larger variants reach up to $\sim$200KB/s. 

\begin{figure}[!tb]
    \centering
    \includegraphics[width=\linewidth]{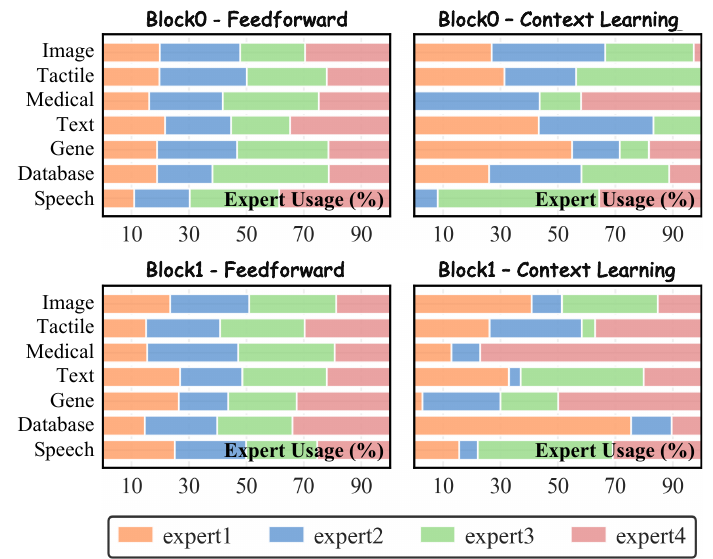}
    \vspace{-18pt}
    \caption{Expert usage of OmniZip-S across blocks and routing modules. Each bar shows the expert usage percent (\%) for a modality in a routing module (context learning or feedforward).}
    \label{fig:routing}
    \vspace{-14pt}
\end{figure}

Besides, we take OmniZip-S as an example to illustrate the routing patterns across modalities and modules, as shown in \cref{fig:routing}. Each bar represents the percent of expert activations for a given modality within a routing module (context learning or feedforward). We observe that context-learning routing exhibits more uneven expert utilization, where some experts may dominate specific modalities. In contrast, feedforward routing shows a more balanced expert usage, suggesting that contextual routing may capture more modality-specific dynamics, whereas feedforward routing tends to promote more generalized representation learning.


\subsection{Ablation Studies}

\begin{table*}[!tb]
    \centering
    \belowrulesep=0pt
    \aboverulesep=0pt
    \renewcommand{\arraystretch}{1.03}
    \setlength{\tabcolsep}{2.7pt}
    \caption{Ablations on all our proposals: modality-unified tokenization, modality-routing contextual learning, modality-routing feedforward, and reparameterization training. Compression performance (bits/Byte) is evaluated on representative dataset from each modality, and the inference speed is measured on a MacBook CPU with a batch size of 128. The chosen configuration is colored in orange.}
    \vspace{-8pt}
    \scriptsize
    \begin{tabular}{cccc|c|ccc|ccccccc}
    \toprule
      
      \multirow{2}{*}{\textbf{\makecell{Unifed\\Tokenizer}}}
      & \multirow{2}{*}{\textbf{\makecell{Contextual\\Routing}}} 
      & \multirow{2}{*}{\textbf{\makecell{Feedforward\\Routing}}} 
      & \multirow{2}{*}{\textbf{Reparam}} 
      & \multirow{2}{*}{\textbf{\makecell{Multi\\Modal}}} 
      & \multirow{2}{*}{\textbf{\#Params}$\downarrow$} 
      & \multirow{2}{*}{\textbf{MACs}$\downarrow$} 
      & \multirow{2}{*}{\textbf{\makecell{Speed\\(KB/s)}}$\uparrow$}
      & \multicolumn{7}{c}{\textbf{bits/Byte$\downarrow$}}\\
      \cmidrule{9-15}

      & & & & & & &
      & \textbf{Kodak}
      & \textbf{TouchandGo}
      & \textbf{Coronal}
      & \textbf{enwik9}
      & \textbf{Spider}
      & \textbf{GenoSeq}
      & \textbf{LibriSpeech}\\
      \midrule
      \ding{55} & \ding{55} & \ding{55} & \ding{55} & \ding{55} 
      & 3.2M & 3.68M & 856 & - & - & - & 1.658 & - & - & - \\
      \ding{55} & \ding{55} & \ding{55} & \checkmark & \ding{55}
      & 3.2M & 3.68M & 856 & - & - & - & 1.590 & - & - & - \\
      \cmidrule{1-15}
      \checkmark & \ding{55} & \ding{55} & \checkmark & \checkmark
      & 3.2M & 3.68M & 856
      & 3.383 & 1.453 & 4.545 & 1.660 & 1.662 & 1.984 & 4.365\\
      \checkmark & \ding{55} & \checkmark & \checkmark & \checkmark 
      & 4.0M & 3.73M & 633
      & 3.352 & 1.431 & 4.314 & 1.424 & 1.448 & 1.915 & 4.295\\
      \checkmark & \checkmark & \ding{55} & \checkmark & \checkmark 
      & 3.0M & 3.82M & 780
      & 3.339 & 1.419 & 4.188 & 1.410 & 1.482 & 1.889 & 4.302\\
      \checkmark & \checkmark & \checkmark & \ding{55} & \checkmark 
      & 4.8M & 3.88M & 500 
      & 3.414 & 1.360 & 4.260 & 1.390 & 1.356 & 1.895 & 4.183\\
      
      \cellcolor{lightpink}\checkmark 
      & \cellcolor{lightpink}\checkmark 
      & \cellcolor{lightpink}\checkmark 
      & \cellcolor{lightpink}\checkmark 
      & \cellcolor{lightpink}\checkmark
      & \cellcolor{lightpink}4.8M 
      & \cellcolor{lightpink}3.88M 
      & \cellcolor{lightpink}500
      & \cellcolor{lightpink}3.307
      & \cellcolor{lightpink}1.338
      & \cellcolor{lightpink}4.179
      & \cellcolor{lightpink}1.370
      & \cellcolor{lightpink}1.320
      & \cellcolor{lightpink}1.868
      & \cellcolor{lightpink}4.155\\
      \bottomrule
    \end{tabular}
    \vspace{-5pt}
    \label{tab:arch-ablations}
\end{table*}

\begin{table*}[!tb]
    \centering
    \belowrulesep=0pt
    \aboverulesep=0pt
    \renewcommand{\arraystretch}{1.03}
    \setlength{\tabcolsep}{4.8pt}
    \caption{Ablations on modality-routing contextual learning. Routing is applied to all blocks with 4 experts and top-$k$=2. We report the total parameters and parameters activated per token. Compression performance (bits/Byte) is evaluated on representative dataset from each modality, and the inference speed is measured on a MacBook CPU with a batch size of 128. The chosen configuration is colored in orange.}
    \vspace{-8pt}
    \scriptsize
    \begin{tabular}{ccc|cc|cc|ccccccc}
    \toprule
      
      \multicolumn{3}{c|}{\textbf{Contextual Routing}}
      & \multicolumn{2}{c|}{\textbf{\#Params}$\downarrow$}  
      & \multirow{2}{*}{\textbf{MACs}$\downarrow$} 
      & \multirow{2}{*}{\textbf{\makecell{Speed (KB/s)}}$\uparrow$}
      & \multicolumn{7}{c}{\textbf{bits/Byte$\downarrow$}}\\
      \cmidrule{1-5}
      \cmidrule{8-14}
      
      \textbf{Receptance} & \textbf{Key} & \textbf{Value}
      & \textbf{Total} & \textbf{Activated} 
      & &
      & \textbf{Kodak}
      & \textbf{TouchandGo}
      & \textbf{Coronal}
      & \textbf{enwik9}
      & \textbf{Spider}
      & \textbf{GenoSeq}
      & \textbf{LibriSpeech}\\
      \midrule

      \checkmark & \ding{55} & \ding{55}
      & 4.8M & 2.7M & 3.88M & 500
      & 3.315 & 1.372 & 4.224 & 1.311 & 1.324 & 1.795 & 4.281\\
      \ding{55} & \checkmark & \ding{55}
      & 4.8M & 2.7M & 3.88M & 500
      & 3.614 & 1.533 & 4.398 & 1.388 & 1.470 & 1.862 & 4.391\\
      
      \cellcolor{lightpink}\ding{55} 
      & \cellcolor{lightpink}\ding{55} 
      & \cellcolor{lightpink}\checkmark
      & \cellcolor{lightpink}4.8M 
      & \cellcolor{lightpink}2.7M 
      & \cellcolor{lightpink}3.88M 
      & \cellcolor{lightpink}500
      & \cellcolor{lightpink}3.307
      & \cellcolor{lightpink}1.338
      & \cellcolor{lightpink}4.179
      & \cellcolor{lightpink}1.370
      & \cellcolor{lightpink}1.320
      & \cellcolor{lightpink}1.868
      & \cellcolor{lightpink}4.155\\
      \cmidrule{1-14}

      \checkmark & \checkmark & \ding{55}
      & 5.4M & 3.0M & 4.02M & 436
      & 3.317 & 1.352 & 4.080 & 1.292 & 1.361 & 1.860 & 4.280\\
      \checkmark & \ding{55} & \checkmark
      & 5.4M & 3.0M & 4.02M & 436
      & 3.313 & 1.360 & 4.140 & 1.297 & 1.309 & 1.813 & 4.178\\
      \ding{55} & \checkmark & \checkmark
      & 5.4M & 3.0M & 4.02M & 436
      & 3.302 & 1.361 & 4.188 & 1.290 & 1.467 & 1.835 & 4.218\\
      \cmidrule{1-14}
      
      \checkmark 
      & \checkmark 
      & \checkmark
      & 6.0M 
      & 3.3M 
      & 4.17M & 382 
      & 3.358 & 1.408 & 4.122 & 1.265 & 1.481 & 1.830 & 4.291\\

      \bottomrule
    \end{tabular}
    \vspace{-5pt}
    \label{tab:moa-ablations}
\end{table*}

\begin{table*}[!tb]
    \centering
    \belowrulesep=0pt
    \aboverulesep=0pt
    \renewcommand{\arraystretch}{1.03}
    \setlength{\tabcolsep}{4.5pt}
    \caption{Ablations on the configurations of the two modality-routing modules (e.g., routed blocks, expert count, top-$k$). We report the total parameters and parameters activated per token. Compression performance (bits/Byte) is evaluated on representative dataset from each modality, and the inference speed is measured on a MacBook CPU with a batch size of 128. The chosen configuration is colored in orange.}
    \vspace{-8pt}
    \scriptsize
    \begin{tabular}{cccc|cc|cc|ccccccc}
    \toprule
      
      \multirow{2}{*}{\textbf{\makecell{Blocks}}}
      & \multirow{2}{*}{\textbf{top-$k$}}  
      & \multirow{2}{*}{\textbf{Experts}} 
      & \multirow{2}{*}{\textbf{\makecell{Hidden\\Factor}}}
      & \multicolumn{2}{c|}{\textbf{\#Params}$\downarrow$}  
      & \multirow{2}{*}{\textbf{MACs}$\downarrow$} 
      & \multirow{2}{*}{\textbf{\makecell{Speed\\(KB/s)}}$\uparrow$}
      & \multicolumn{7}{c}{\textbf{bits/Byte$\downarrow$}}\\
      \cmidrule{5-6}
      \cmidrule{9-15}

      & & &
      & \textbf{Total} & \textbf{Activated} 
      & &
      & \textbf{Kodak}
      & \textbf{TouchandGo}
      & \textbf{Coronal}
      & \textbf{enwik9}
      & \textbf{Spider}
      & \textbf{GenoSeq}
      & \textbf{LibriSpeech}\\
      \midrule

      1 & 1 & 4 & $2\times$ & 3.6M & 2.2M & 3.65M & 594 
      & 3.436 & 1.150 & 4.267 & 1.391 & 1.511 & 1.878 & 4.336\\
      1 & 2 & 4 & $2\times$ & 3.6M & 2.5M & 3.85M & 570 
      & 3.358 & 1.420 & 4.230 & 1.380 & 1.344 & 1.896 & 4.307\\
      
      \cellcolor{lightpink}2 
      & \cellcolor{lightpink}2 
      & \cellcolor{lightpink}4 
      & \cellcolor{lightpink}$2\times$
      & \cellcolor{lightpink}4.8M 
      & \cellcolor{lightpink}2.7M
      & \cellcolor{lightpink}3.88M
      & \cellcolor{lightpink}500
      & \cellcolor{lightpink}3.307
      & \cellcolor{lightpink}1.338
      & \cellcolor{lightpink}4.179
      & \cellcolor{lightpink}1.370
      & \cellcolor{lightpink}1.320
      & \cellcolor{lightpink}1.868
      & \cellcolor{lightpink}4.155\\
      
      2 & 2 & 3 & $2\times$ & 4.0M & 2.7M & 3.88M & 500 
      & 3.355 & 1.424 & 4.154 & 1.369 & 1.410 & 1.882 & 4.300\\
      2 & 2 & 5 & $2\times$ & 5.2M & 2.7M & 3.88M & 500 
      & 3.232 & 1.416 & 4.190 & 1.358 & 1.352 & 1.855 & 4.206\\
      \cmidrule{1-15}
      
      2 & 2 & 4 & $1\times$ & 3.2M & 2.0M & 3.48M & 533 
      & 3.502 & 1.560 & 4.216 & 1.392 & 1.590 & 1.875 & 4.380\\
      
      \cellcolor{lightpink}2 
      & \cellcolor{lightpink}2 
      & \cellcolor{lightpink}4 
      & \cellcolor{lightpink}$2\times$
      & \cellcolor{lightpink}4.8M 
      & \cellcolor{lightpink}2.7M
      & \cellcolor{lightpink}3.88M
      & \cellcolor{lightpink}500
      & \cellcolor{lightpink}3.307
      & \cellcolor{lightpink}1.338
      & \cellcolor{lightpink}4.179
      & \cellcolor{lightpink}1.370
      & \cellcolor{lightpink}1.320
      & \cellcolor{lightpink}1.868
      & \cellcolor{lightpink}4.155\\
      
      2 & 2 & 4 & $4\times$ & 8.0M & 4.4M & 4.65M & 325 
      & 3.316 & 1.354 & 4.110 & 1.331 & 1.315 & 1.850 & 4.157\\
      \bottomrule
    \end{tabular}
    \vspace{-11pt}
    \label{tab:moe-ablations}
\end{table*}

\paragraph{Discussion on Multi-Modal Proposals.}


\cref{tab:arch-ablations} reports ablations of our multi-modal components: the unified tokenizer enables multi-modality, while contextual and feedforward routing provide modality adaptivity in context modeling and nonlinear transformation. We run all ablations on OmniZip-S with the same hyperparameters as the main experiments. It can be seen that removing either the routing mechanism leads to a clear drop in multi-modal compression performance, confirming their effectiveness in enhancing multi-modal compression.



\vspace{-11pt}
\paragraph{Discussion on Modality-Routing Configurations.}

We analyze the configurations of the two modality-routing modules in contextual learning and feedforward stages. 

For contextual learning, we apply routing to different projection layers with 4 experts and top-$k$ of 2. As shown in \cref{tab:moa-ablations}, applying routing to the V projection yields better compression performance than applying it to the K/R projections. Extending routing to more projections slightly increases model capacity but also leads to 10\%$\sim$20\% more parameters. Hence, we apply routing solely to the V projection in each contextual learning stage.

Then, we vary the number of routed blocks (from half to all), expert count (from 3 to 5), and top-$k$ (from 1 to 2) for the contextual and feedforward routing modules, as shown in \cref{tab:moe-ablations}. It can be seen that using 4 experts and top-$k$ of 2 yields the best trade-off between complexity and performance. We also vary the hidden factor of the feedforward routing module’s MLP experts. Moreover, setting the feedforward experts with a hidden factor of $2\times$ strikes a good balance between compression performance and model compactness, hence it is adopted for OmniZip models.



\section{Conclusion}

We present OmniZip, a unified and lightweight lossless compressor designed for multi-modal data. 
It supports most real-world data types suitable for lossless compression: image-like data (e.g., natural image, medical image, tactile signals), text-like data (e.g., natural language, gene sequence, database), and speech. 
Built on a lightweight backbone, OmniZip incorporates three components to enable efficient multi-modal compression: modality-unified tokenization, modality-routing context learning, and modality-routing feedforward. A reparameterization training strategy is also used to improve model capacity. OmniZip achieves compression performance on par with or better than existing state-of-the-art methods across multiple modalities, while using much fewer parameters. It also supports near real-time inference even on resource-constrained edge devices, like MacBook CPUs and iPhone NPUs.

\section{Acknowledgments}

This work was partly supported by the NSFC (62431015, 62571317, 62501387), the Fundamental Research Funds for the Central Universities, Shanghai Key Laboratory of Digital Media Processing and Transmission under Grant 22DZ2229005, 111 project BP0719010 and the Okawa Research Fund, the Ant Group Research Fund.
{
    \small
    \normalem
    \bibliographystyle{ieeenat_fullname}
    \bibliography{main}
}

\clearpage
\setcounter{page}{1}
\maketitlesupplementary

The supplementary material includes deeper explanations of the proposed method and extended experiments, covering: (1) detailed descriptions of the modality-unified tokenization, modality-routing context learning, modality-routing feedforward design, and reparameterization training strategy; (2) implementation details of multi-modal datasets and training/testing process; (3) extended lossless compression performance using adjusted bits/Byte as the metric.


\section{Backbone Selection}

We compare three common backbone architectures for the probability prediction model: Transformer \cite{transformer:2017}, Mamba \cite{mamba}, and RWKV \cite{rwkv7}. These models are evaluated in terms of compression efficiency (bits/Byte), computational cost (MACs), and inference speed (KB/s). 

As for Transformer, We use a causal, decoder-only Transformer implemented in JAX/Haiku. Input tokens are embedded, scaled, and summed with sinusoidal positional encodings, then processed through a stack of eight multi-head self-attention (8 heads) and feed-forward layers, each followed by residual connections and LayerNorm. For Mamba and RWKV, we adopt their official implementations\footnote{Mamba: \url{https://github.com/state-spaces/mamba}.}\footnote{RWKV: \url{https://github.com/BlinkDL/RWKV-LM}.}. The trained models are converted into CoreML packages and benchmarked on a MacBook Pro 2024 with an Apple M4 chip using CPU inference. 


\section{Modality-Unified Tokenization}
\label{sec:tokenization}

To enable multi-modal lossless compression, we design a reversible modality-unified tokenization strategy that maps heterogeneous data into a shared token space. The main challenge arises from the fundamentally distinct structures of different modalities. Moreover, to ensure lossless reconstruction, the tokenization process must be fully invertible.

For text-like data, including natural language, gene sequences, and databases,
we follow \cite{l3tc} and adopt a SentencePiece BPE tokenizer \cite{bpe} with a vocabulary size of 16K. To better adapt to domain-specific characteristics, we extend the vocabulary with symbolic tokens: like nucleotide bases (A, T, G, C) for gene sequences, and common SQL keywords (e.g., SELECT, FROM, WHERE) for databases. 
Formally, given a text string $S$, the tokenizer performs:
\begin{equation}
S_{\text{text}} \xrightarrow{\text{SPM BPE}} [x_1, x_2, \dots, x_n],\ \ x_i \in \mathcal{V}_{\text{text}},\ |\mathcal{V}_{\text{text}}|=16\text{K}.
\end{equation}

For image-like data, including natural, medical, and tactile images,
we partition each image into $16\times16\times3$ patches to preserve local spatial correlations. Pixels within each patch are flattened in raster-scan order, and each RGB channel is expanded sequentially: $[R_1, G_1, B_1, R_2, G_2, B_2, \dots]$. We treat each 8-bit sub-pixel as an independent token (i.e., $|\mathcal{V}_{\text{image}}| = 256$), thereby maintaining inter-pixel and inter-channel correlations. For grayscale medical images, each 8-bit intensity value is directly mapped to a single token. Tactile data are handled in two ways: visual tactile maps are processed as regular RGB images, while 3D tactile force vectors $(x, y, z)$ are linearly mapped into pseudo-RGB triplets for tokenization.

Speech is continuous and hard to discretize without loss. Hence, we adopt a next-byte prediction scheme, reading the raw byte stream and treating each byte as a token:
\begin{equation}
S_{\text{speech}}(t) \xrightarrow{\text{each byte}} [x_1, x_2, \dots, x_n], \quad 
x_i \in [0, 255],
\end{equation}
yielding a 256-size vocabulary shared with image-like modalities (i.e., $|\mathcal{V}_{\text{speech}}| = 256$). 

After tokenization, we merge all modality vocabularies into a single unified token space:
\begin{equation}
\mathcal{V}_{\text{uni}} = \mathcal{V}_{\text{text}} \cup \mathcal{V}_{\text{image}} \cup \mathcal{V}_{\text{speech}}.
\end{equation}
Each sequence is prefixed with a modality identifier token, i.e., $<$image$>$, $<$medical$>$, $<$tactile$>$, $<$text$>$, $<$gene$>$, $<$database$>$, or $<$speech$>$, allowing the model to learn modality-aware priors during probability estimation. 

Before softmax and arithmetic coding, we apply modality-specific masking to restrict the active token propabilities and improve the compression efficiency. For example, during image compression, only tokens in $\mathcal{V}_{\text{image}}$ are retained, and all others are zeroed out:
\begin{equation}
p_{\text{image}}(x_i|x_{<i}) = \text{softmax}(o(x_i|x_{<i}) \odot M_{\text{image}}),
\end{equation}
where $M_{\text{image}}$ is a binary mask selecting image tokens, $o(x_i|x_{<i}$ is the output logit before softmax.

\section{Modality-Routing Context Learning}
\label{supp:moa}

\begin{figure}
    \centering
    \includegraphics[width=\linewidth]{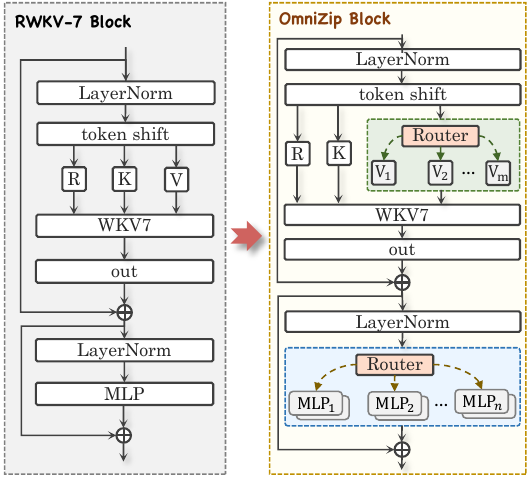}
    \caption{Key structural differences between RWKV-7 and OmniZip model. Left: an RWKV-7 block. Right: an OmniZip block.}
    \label{fig:structure-supp}
\end{figure}

As shown in \cref{fig:structure-supp}, each RWKV block \cite{rwkv7} consists of two components: a Time Mixing module and a multilayer perceptron (MLP) \cite{mlp}. The Time Mixing module serves as a lightweight alternative to attention, modeling contextual dependencies through a recurrent formulation. Instead of computing pairwise attention, it maintains a running state that summarizes past tokens, allowing linear-time inference.

As stated, different modalities exhibit distinct contextual dependencies. For instance, text are one-dimensional sequential and have mostly short-range semantic dependencies. Images are two-dimensional and flattened into sequences, with inter-pixel and inter-channel correlations. Speech is continuous and temporally structured. 

To enhance multi-modal adaptability, we integrate a mixture-of-experts (MoE) mechanism \cite{moe} into the Time Mixing module. MoE dynamically selects specialized experts based on input tokens, enabling conditional computation and reducing unnecessary activation. Specifically, given an input token $x_i$, a router network predicts a score:
\begin{equation}
g_{i,e} = \text{softmax}(x_i\cdot W_g)_e =
\frac{\exp(x_i\cdot W_{g,e})}{\sum_{e'=1}^{E} \exp(x_i\cdot W_{g,e'})},
\end{equation}
where $E$ is the expert count, $W_g$ is the routing projection.
$g_{i,e}$ means the likelihood that expert $e$ should process $x_i$.

To keep efficiency, we adopt top-$k$ sparse routing, where only the experts with top-$k$ highest scores are activated, and their outputs are aggregated as a weighted sum. 
We experiment with applying MoE to different components and find that the V layer benefits most from such adaptive routing. Conceptually, the K layer serves as a semantic index, and the R layer acts as a temporal gate, while the V layer encodes concrete contextual content. Allowing diversity in V layer enables the model to adapt more flexibly to modality-specific information. Hence, we apply MoE only to the V layer, sharing K and R layers across all V experts. Formally, the V-projection output in our MoE design is computed as:
\begin{equation}
\text{V}(x_i) = \sum_{e\in \text{top-}k} \hat{g}_{i,e} , e(x_i),
\end{equation}
where each expert $e$ is a distinct V projection layer, and $\hat{g}_{i,e}$ denotes the re-normalized routing score. 

To ensure model compactness, we use $E=4$ experts and $k=2$. This setup increases the number of parameters by only three additional V layers per Time Mixing module, yet notably improves the multi-modal adaptability.

\section{Modality-Routing Feedforward}

In the RWKV backbone \cite{rwkv7}, each block integrates a feedforward MLP that projects intermediate representations into a higher-dimensional space, applies nonlinear transformations, and fuses contextual information from the Time Mixing module. This structure enhances model expressiveness but treats all tokens uniformly, regardless of their modality. In fact, a single shared MLP is insufficient to capture the distinct properties of heterogeneous modalities.

To better support multi-modal compression, we replace this MLP with a modality-routing feedforward module based on MoE. The routing mechanism follows the design in \cref{supp:moa}, while the experts here are small MLPs instead of V layers. Each MLP expert adopts a hidden factor of $2\times$, half that of the original large MLP. With four experts and top-$k=2$, the number of activated parameters per token remains nearly identical to the original design. 

\begin{figure}[!tb]
    \centering
    \includegraphics[width=\linewidth]{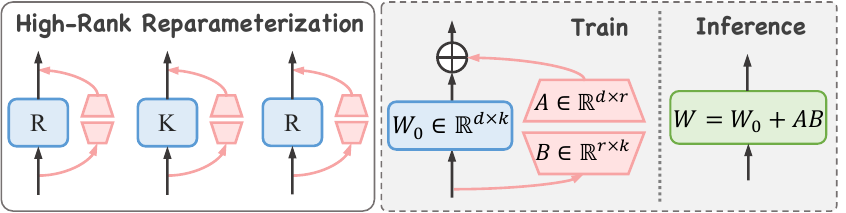}
    \caption{Illustration of the reparameterization training strategy.}
    \label{fig:reparam}
\end{figure}

\section{Reparameterization Training Strategy}

Following \cite{l3tc}, we adopt a high-rank reparameterization training strategy to enhance OmniZip’s compression capacity without increasing inference complexity. During training, each R, K, and V projection layer in the Time Mixing module is augmented with an auxiliary high-rank branch to improve representational power. At inference, these auxiliary branches are merged back into the main path through structural reparameterization, maintaining a compact single-path design, as illustrated in \cref{fig:reparam}.

\begin{table*}[!tb]
    \belowrulesep=0pt
    \aboverulesep=0pt
    \centering
    \renewcommand{\arraystretch}{1.25}
    \setlength{\tabcolsep}{2.6pt}
    \newcommand{\centerdash}[1]{\ifx#1-\multicolumn{1}{c}{-}\else#1\fi}
    \caption{Details of the utilized multi-modal datasets (16 datasets from 7 modalities).} 
    \footnotesize   
    \vspace{-4pt}
    \begin{tabular}{llccl}
    \toprule
    \textbf{Dataset}  
    & \textbf{Type} 
    & \textbf{Train} 
    & \textbf{Test} 
    & \textbf{Description} \\
    \midrule
    Kodak \cite{kodak} 
    & image & \ding{55} & \checkmark
    & 24 film-scanned photos ($768\times512$) with balanced natural and synthetic scenes. We use it for testing. \\
    CLIC-P \cite{clic} 
    & image & \ding{55} & \checkmark
    & 41 high-resolution DSLR images (mostly 2K) with rich color and texture diversity. We use it for testing.\\
    CLIC-M \cite{clic} 
    & image & \ding{55} & \checkmark
    & 61 smartphone photos (mostly 2K) containing real-world noise and artifacts. We use it for testing.\\
    DIV2K \cite{div2k}
    & image & \checkmark & \checkmark
    & 900 2K-resolution images with a large diversity of contents. We follow the official split (800 train, 100 test).\\
    \midrule

    Axial \cite{mrnet}
    & medical & \checkmark & \checkmark
    & 2D knee MRI slices ($256\times256$) in the axial plane. We follow the official split ($\sim$38K train, $\sim$4.1K test).\\
    Coronal \cite{mrnet} 
    & medical & \checkmark & \checkmark
    & 2D knee MRI slices ($256\times256$) in the coronal plane. We follow the official split ($\sim$33K train, $\sim$3.5K test).\\
    Sagittal \cite{mrnet}
    & medical & \checkmark & \checkmark
    & 2D knee MRI slices ($256\times256$) in the sagittal plane. We follow the official split ($\sim$34K train, $\sim$3.6K test).\\
    \midrule

    TouchandGo \cite{touchandgo}
    & tactile & \checkmark & \checkmark
    & A tactile dataset of 140 trajectories ($640\times480$), with 60\% for training and the rest for testing. \\
    ObjectFolder \cite{objectfolder}
    & tactile & \checkmark & \checkmark
    & A tactile dataset of 1000 trajectories ($120\times160$), with 60\% for training and the rest for testing. \\
    \midrule
    
    enwik8 \cite{enwik8}
    & text & \checkmark & \ding{55}
    & The first 100MB of the English Wikipedia dump. We use it for testing.\\
    enwik9 \cite{enwik9}
    & text & \ding{55} & \checkmark
    & The first 1GB of the English Wikipedia dump. We use it for testing. \\
    Gutenberg \cite{gutenberg}
    & text & \checkmark & \checkmark
    & A library containing 75K eBooks. We randomly select 2K and 1K books for training and testing, respectively.\\
    \midrule
    
    GenoSeq \cite{genoseq} 
    & gene & \checkmark & \checkmark
    & 17 genomics sequencing dataset with FastQ format. We use 10 for training and 7 for testing.\\
    
    DNACorpus \cite{dnacorpus} 
    & gene & \checkmark & \checkmark
    & DNA sequences of 15 species, we follow the official split (370MB for training and 246MB for testing).\\
    \midrule

    Spider \cite{spider} 
    & database & \checkmark & \checkmark 
    & 5693 SQL queries on 200 databases. We follow the official split (314MB for training and 60MB for testing).\\
    
    WikiSQL \cite{wikisql} 
    & database & \checkmark & \checkmark 
    & 24241 SQL queries from Wikipedia. We follow the official split (38MB for training and 12MB for testing).\\
    \midrule

    LibriSpeech \cite{librispeech} 
    & speech & \checkmark & \checkmark
    & A 1000-hour English speech dataset. We follow the official split (1GB for training and 600MB for testing).\\
    
    \bottomrule
    \end{tabular}
    \vspace{-3pt}
    \label{tab:datasets}
\end{table*}

Specifically, instead of using parallel branches like $1 \times 1$ convolutions or shortcuts as in \cite{reparam}, we follow the strategy in \cite{l3tc} and reparameterize each branch as a product of two high-rank matrices: $A \in \mathbb{R}^{d \times r}$ and $B \in \mathbb{R}^{r \times k}$, where the main-path's weight is $W_0 \in \mathbb{R}^{d \times k}$ and $r \gg d, k$. During training, the layer's output is the sum of the main branch and the bypass branch. These branches are merged at inference via structural reparameterization: $W = W_0 + A \times B$, resulting in a single-path structure that reduces both runtime and memory usage. Importantly, although the branches are merged during inference, the learned multi-branch parameters are preserved, maintaining high capability. 

In OmniZip, this strategy is applied to all the R/K/V layers in the Time Mixing module. Following \cite{l3tc}, we set the decomposition rank $r$ to $4\times$ the embedding dimension.


\section{Multi-Modal Datasets}

We conduct experiments across seven data types, including natural images, medical images, tactile signals, text, gene sequences, databases, and speech, covering a total of 16 datasets for training and evaluation, as shown in \cref{tab:datasets}. 

To ensure balanced and stable multi-modal training, we normalize the data scale across modalities by restricting each modality’s training set to approximately 1 GB. For large datasets, samples are randomly drawn to maintain representative coverage of content diversity, while smaller datasets are augmented through simple transformations such as random cropping/flipping, token shuffling, or noise perturbation (depending on the modality type).

During training, we also adopt a balanced batch sampling strategy to prevent overfitting toward some modalities. Specifically, the dataloader cycles through all modalities in a round-robin manner, dynamically constructing each batch by uniformly sampling from the shuffled indices of every modality. This ensures that each batch contributes an equal number of samples per modality, allowing the model to learn modality-invariant patterns.

Regarding the model input format, as stated in \cref{sec:tokenization}, for image-like data, each input corresponds to a flattened $16\times16\times3$ patch, resulting in a sequence length of 768 tokens. Text-like and speech data are processed as sequential streams, with each input sequence containing 1024 tokens.




\section{Training and Testing Process}

To accelerate the overall pipeline, we incorporate optimizations like Cython \cite{cython}, Numba \cite{numba}, and distributed parallelization. For model evaluation and deployment, we test OmniZip across three hardware platforms: (1) NVIDIA A100 GPU (for training and high-throughput testing), (2) MacBook Pro 2024 (Apple M4 CPU), and (3) iPhone 17 Pro (A19 NPU). For CPU and NPU evaluation, we convert the trained PyTorch models into CoreML packages \cite{coreml} and conduct profiling via Xcode Instruments. For GPU evaluation, we switch the GPU to performance mode and ensure no other processes are running on the device. We then measure inference latency by executing the model 1000 times consecutively and report the average runtime.
 


\section{Adjusted Compression Results}

To provide a more realistic evaluation of deployment efficiency, we adopt the adjusted bits-per-Byte metric, which incorporates model size into the compressed data size. It penalizes large models and reflects the trade-off between compression performance and model storage overhead.


\begin{table*}[tb]
    \belowrulesep=0pt
    \aboverulesep=0pt
    \centering
    \footnotesize
    \renewcommand{\arraystretch}{1.1}
    \setlength{\tabcolsep}{4.5pt}
    \newcommand{\centerdash}[1]{\ifx#1-\multicolumn{1}{c}{-}\else#1\fi}
    \caption{Adjusted compression performance (bits/Byte) on image-like (natual image, medical image, and tactile) datasets. $*$ denotes pretrained LLMs, $\dag$ indicates that some results are reprodeced by us. Other values are taken from their papers.} 
    \vspace{-4pt}
    \begin{tabular}
    {p{1.7cm}|l|c|llll|ll|lll}
    \toprule
     \multirow{2}{*}{\textbf{Compressor}} 
     & \multirow{2}{*}{\textbf{\#Params}$\downarrow$} 
     & \multirow{2}{*}{\textbf{\makecell{Multi\\Modal}}} 
     & \multicolumn{9}{c}{\textbf{Adjusted bits/Byte$\downarrow$}}\\[1pt]
     \cmidrule{4-12}

     & & 
     & \textbf{Kodak} & \textbf{CLIC-P} & \textbf{CLIC-M} & \textbf{DIV2K} 
     & \textbf{TouchandGo} & \textbf{ObjectFolder} 
     & \textbf{Axial} & \textbf{Coronal} & \textbf{Sagittal}\\
    \midrule

    \cellcolor{lightgrey}$\text{Llama3}^{*\dag}$ \cite{lmic} 
    & \cellcolor{lightgrey}8B 
    & \cellcolor{lightgrey}\checkmark 
    & \cellcolor{lightgrey}8490321 
    & \cellcolor{lightgrey}971320
    & \cellcolor{lightgrey}553618
    & \cellcolor{lightgrey}291465
    & \cellcolor{lightgrey}322642
    & \cellcolor{lightgrey}214197
    & \cellcolor{lightgrey}693352
    & \cellcolor{lightgrey}750368
    & \cellcolor{lightgrey}970961 \\
    
    \cellcolor{lightgrey}$\text{RWKV}^{*\dag}$ \cite{lmic}  
    & \cellcolor{lightgrey}7B 
    & \cellcolor{lightgrey}\checkmark 
    & \cellcolor{lightgrey}7429031 
    & \cellcolor{lightgrey}849906
    & \cellcolor{lightgrey}484417
    & \cellcolor{lightgrey}255032
    & \cellcolor{lightgrey}282312
    & \cellcolor{lightgrey}187423
    & \cellcolor{lightgrey}606683
    & \cellcolor{lightgrey}656573
    & \cellcolor{lightgrey}823341\\
    
    DLPR$^{*\dag}$ \cite{dlpr}   & 22.3M 
    & \ding{55} 
    & 23670 & 2710 & 1545 & 815 
    & 900   & 601 
    & - & - & - \\
    
    L3C$^{*\dag}$ \cite{l3c}     & 5M 
    & \ding{55} 
    & 5310  & 610 & 349 & 185 
    & 203   & 138 
    & 439   & 473 & 594 \\
    
    P2LLM \cite{p2llm} & 8B 
    & \ding{55} 
    & 8490319
    & 971318
    & 553616
    & 291463
    & - & - & - & - & -\\

    aiWave \cite{aiwave} & 695M 
    & \ding{55}
    & - & - & - & -
    & - & - 
    & 59372 & 64254 & 80574\\
    
    BCM \cite{bcm} & 12.5M 
    & \ding{55}
    & - & - & - & -
    & - & - 
    & 1088 & 1176 & 1475\\
    
    \midrule

    \cellcolor{lightpink}OmniZip-S   
    & \cellcolor{lightpink}4.8M 
    & \cellcolor{lightpink}\checkmark 
    & \cellcolor{lightpink}5097  
    & \cellcolor{lightpink}586
    & \cellcolor{lightpink}335
    & \cellcolor{lightpink}178 
    & \cellcolor{lightpink}195
    & \cellcolor{lightpink}132
    & \cellcolor{lightpink}421
    & \cellcolor{lightpink}454
    & \cellcolor{lightpink}570\\
    
    \cellcolor{lightpink}OmniZip-M   
    & \cellcolor{lightpink}38M 
    & \cellcolor{lightpink}\checkmark 
    & \cellcolor{lightpink}40332  
    & \cellcolor{lightpink}4618
    & \cellcolor{lightpink}2632
    & \cellcolor{lightpink}1387
    & \cellcolor{lightpink}1534
    & \cellcolor{lightpink}1020
    & \cellcolor{lightpink}3298
    & \cellcolor{lightpink}3568
    & \cellcolor{lightpink}4475\\
    
    \cellcolor{lightpink}OmniZip-L   
    & \cellcolor{lightpink}152M 
    & \cellcolor{lightpink}\checkmark 
    & \cellcolor{lightpink}161319  
    & \cellcolor{lightpink}18458
    & \cellcolor{lightpink}10521
    & \cellcolor{lightpink}5540
    & \cellcolor{lightpink}6131
    & \cellcolor{lightpink}4072
    & \cellcolor{lightpink}13178
    & \cellcolor{lightpink}14261
    & \cellcolor{lightpink}17883\\
    
    \bottomrule
    \end{tabular}
    \label{tab:compare-image-tactile-medical-supp}
\end{table*}

\begin{table*}[tb]
    \belowrulesep=0pt
    \aboverulesep=0pt
    \centering
    \renewcommand{\arraystretch}{1.1}
    \setlength{\tabcolsep}{7.7pt}
    \newcommand{\centerdash}[1]{\ifx#1-\multicolumn{1}{c}{-}\else#1\fi}
    \caption{Adjusted compression performance (bits/Byte) on text-like (natural language, gene sequence, database) and speech datasets. $*$ denotes pretrained LLMs, $\dag$ indicates that some results are reprodeced by us. Other values are taken from their papers.} 
    \footnotesize   
    \vspace{-4pt}
    \begin{tabular}
    {p{1.7cm}|l|c|ll|ll|ll|l}
    \toprule
     \multirow{2}{*}{\textbf{Compressor}} 
     & \multirow{2}{*}{\textbf{\#Params}$\downarrow$} 
     & \multirow{2}{*}{\textbf{\makecell{Multi\\Modal}}} 
     & \multicolumn{7}{c}{\textbf{Adjusted bits/Byte$\downarrow$}}\\[1pt]
     \cmidrule{4-10}

     & & 
     & \textbf{enwik9}  & \textbf{Gutenberg} 
     & \textbf{Spider}  & \textbf{WikiSQL} 
     & \textbf{GenoSeq} & \textbf{DNACorpus} 
     & \textbf{LibriSpeech} \\
    \midrule

    \cellcolor{lightgrey}$\text{Llama3}^{*\dag}$ \cite{lmic} 
    & \cellcolor{lightgrey}8B 
    & \cellcolor{lightgrey}\checkmark 
    & \cellcolor{lightgrey}129 
    & \cellcolor{lightgrey}362459
    & \cellcolor{lightgrey}2125
    & \cellcolor{lightgrey}10835
    & \cellcolor{lightgrey}500
    & \cellcolor{lightgrey}1138
    & \cellcolor{lightgrey}212 \\
    
    \cellcolor{lightgrey}$\text{RWKV}^{*\dag}$ \cite{lmic}  
    & \cellcolor{lightgrey}7B 
    & \cellcolor{lightgrey}\checkmark 
    & \cellcolor{lightgrey}113 
    & \cellcolor{lightgrey}317152
    & \cellcolor{lightgrey}1859
    & \cellcolor{lightgrey}9480
    & \cellcolor{lightgrey}437
    & \cellcolor{lightgrey}996
    & \cellcolor{lightgrey}186\\

    tszip \cite{tszip} & 169 & \ding{55}
    & 4 & 7658
    & 46 & 230
    & 13 & 26
    & - \\

    L3TC \cite{l3tc} & 12 & \ding{55}
    & 2 & 546
    & 6 & 18
    & 3 & 4
    & - \\

    \midrule

    \cellcolor{lightpink}OmniZip-S   
    & \cellcolor{lightpink}4.8M 
    & \cellcolor{lightpink}\checkmark 
    & \cellcolor{lightpink}1  
    & \cellcolor{lightpink}219 
    & \cellcolor{lightpink}3
    & \cellcolor{lightpink}8
    & \cellcolor{lightpink}2
    & \cellcolor{lightpink}3
    & \cellcolor{lightpink}4\\
    
    \cellcolor{lightpink}OmniZip-M   
    & \cellcolor{lightpink}38M 
    & \cellcolor{lightpink}\checkmark 
    & \cellcolor{lightpink}2  
    & \cellcolor{lightpink}1723
    & \cellcolor{lightpink}11
    & \cellcolor{lightpink}52
    & \cellcolor{lightpink}4
    & \cellcolor{lightpink}7
    & \cellcolor{lightpink}5\\
    
    \cellcolor{lightpink}OmniZip-L   
    & \cellcolor{lightpink}152M 
    & \cellcolor{lightpink}\checkmark 
    & \cellcolor{lightpink}3
    & \cellcolor{lightpink}6888  
    & \cellcolor{lightpink}42
    & \cellcolor{lightpink}207
    & \cellcolor{lightpink}11
    & \cellcolor{lightpink}23
    & \cellcolor{lightpink}8\\
    
    \bottomrule
    \end{tabular}
    \vspace{-5pt}
    \label{tab:compare-text-gene-database-speech-supp}
\end{table*}

As shown in \cref{tab:compare-image-tactile-medical-supp} and \cref{tab:compare-text-gene-database-speech-supp}, OmniZip-S achieves superior adjusted compression efficiency across all seven modalities, including image-like, text-like, and speech datasets. OmniZip-M and OmniZip-L also exhibit satisfactory balance between compression efficiency and deployability. Among modality-specific compressors, \citet{l3c} and \citet{l3tc} achieve competitive adjusted compression efficiency on image-like and text-like datasets, respectively. Though \cite{lmic} demonstrate strong compression efficiency on text-like and speech datasets, their massive parameter sizes dominate the total compression cost, resulting in significantly higher adjusted bits-per-Byte values.


\end{document}